\documentclass[journal]{IEEEtran}
\usepackage{amsmath}
\usepackage{graphicx}
\usepackage{xcolor}
\usepackage{booktabs}
\usepackage{rotating} 
\usepackage{booktabs}
 \usepackage{algorithm}
 \usepackage{algorithmic}

\usepackage[colorinlistoftodos]{todonotes}
\usepackage{lipsum}  
\usepackage{cite} 
\usepackage{authblk} 
\usepackage[margin=1in]{geometry} 

\usepackage{cuted}

\ifCLASSINFOpdf
\else
\fi
\begin{document}
%
\title{Integrating LLM, EEG, and Eye-Tracking Biomarker Analysis for Word-Level Neural State Classification in Semantic Inference Reading Comprehension}
%
%
%

\author{Yuhong Zhang, Qin Li,
        Sujal Nahata, Tasnia Jamal, Shih-kuen Cheng, Gert Cauwenberghs,~\IEEEmembership{Fellow,~IEEE}, Tzyy-Ping Jung,~\IEEEmembership{Fellow,~IEEE}
\thanks{Yuhong Zhang is with the School of Engineering, Brown University, Providence,
RI, 02912, USA and Institute for Neural Computation, University of California San Diego, La Jolla, CA, 92093, USA and Department of Radiology and Biomedical Imaging, Yale University, New Haven, CT, 06520, USA, e-mail: yuhong\_zhang1@brown.edu}
\thanks{Qin Li is with the Department of Bioengineering, University of California Los Angeles, Los Angeles, CA, 90095, USA, e-mail: qinli2021@g.ucla.edu }
\thanks{Sujal Nahata is with the Department of Computer Science and Engineering, University of California San Diego, La Jolla, CA, 92093, USA e-mail: snahata@ucsd.edu }
\thanks{Tasnia Jamal is with the Department of Electrical and Computer Engineering, University of California San Diego, La Jolla, CA, 92093, USA e-mail: tjamal@ucsd.edu }
\thanks{Shih-kuen Cheng is with the Institute of Cognitive Neuroscience, National Central University, Taoyuan, 32001, Taiwan e-mail: Skcheng@cc.ncu.edu.tw}
\thanks{Gert Cauwenberghs is with the Department of Bioengineering and Institute for Neural Computation, University of California San Diego, La Jolla, CA, 92093 USA e-mail: gcauwenberghs@ucsd.edu }
\thanks{Tzyy-Ping Jung is with the Institute for Neural Computation, University of California San Diego, La Jolla, CA, 92093 USA, e-mail: tpjung@ucsd.edu }
\thanks{Manuscript received XXX XX, 2023; revised XXX XX, XXX.}}

%
%

\markboth{}
{Shell \MakeLowercase{\textit{et al.}}: Bare Demo of IEEEtran.cls for IEEE Journals}
%



\maketitle

\begin{abstract}
With the recent proliferation of large language models (LLMs), such as Generative Pre-trained Transformers (GPT), there has been a significant shift in exploring human and machine comprehension of semantic language meaning. This shift calls for interdisciplinary research that bridges cognitive science and natural language processing (NLP). This pilot study aims to provide insights into individuals' neural states during a semantic relation reading-comprehension task. We propose jointly analyzing LLMs, eye-gaze, and electroencephalographic (EEG) data to study how the brain processes words with varying degrees of relevance to a keyword during reading. We also use a feature engineering approach to improve the fixation-related EEG data classification while participants read words with high versus low relevance to the keyword. The best validation accuracy in this word-level classification is over 60\% across 12 subjects. Words of high relevance to the inference keyword had significantly more eye fixations per word: 1.0584 compared to 0.6576 when excluding no-fixation words, and 1.5126 compared to 1.4026 when including them. This study represents the first attempt to classify brain states at a word level using LLM knowledge. It provides valuable insights into human cognitive abilities and the realm of Artificial General Intelligence (AGI), and offers guidance for developing potential reading-assisted technologies.
\end{abstract}


%
\IEEEpeerreviewmaketitle

\section{Introduction}
%
%
%
%
\IEEEPARstart{R}{ecent} advancements in LLMs and generative AI have significantly impacted various aspects of human society and industry. Notable examples include GPT-X models developed by OpenAI and Midjourney, among others \cite{wang2023scientific, singhal2023large, abdullah2022chatgpt, bubeck2023sparks}. As artificial agents improve their proficiency, it becomes increasingly crucial to deepen our understanding of machine learning, decision-making processes, and human cognitive functions \cite{gunning2019xai}. For instance, both humans and machines employ strategies for semantic inference. Humans extract crucial information from texts via specific gaze patterns during reading \cite{just1980theory,rayner1998eye,kintsch1998comprehension}, whereas language models predict subsequent words using contextual cues \cite{binz2023using}. Therefore, this pilot study raises the question: Can we differentiate individuals' mental states when their gaze fixates on words of varying significance within a sentence, particularly at a word level, during tasks involving semantic inference and reading comprehension? 

The successfulness of the prediction tasks could have significant implications for current machine learning applications and both science and technology, such as Human-in-the-loop Machine Learning \cite{ouyang2022training}, Brain-Computer Interfaces (BCI) for text communications \cite{pandarinath2017high}, and personalized Learning and Accessibility Tools in real-time \cite{shawky2019towards}.

Previous studies demonstrate biomarkers that affirm patterns in subjects during reading comprehension tasks. For example, several neurobiological markers linked to reading comprehension, including P300 and N400, were first identified in the 1980s \cite{kutas2011thirty}. As the groundbreaking research in reading comprehension, the study revealed that there are distinct patterns in N400 for “semantic moderate” and “semantic strong” words \cite{kutas1980reading}. 

Furthermore, numerous classical theories within the cognitive science community aim to elucidate and delineate the processes through which humans comprehend text and make inferences. Kintsch \cite{kintsch1988role} introduced the Construction-Integration (CI) model, which posits text comprehension as a two-stage process: initially constructing a textbase (comprehending the text at the surface and propositional level) and subsequently integrating it with prior knowledge to form a situation model (a mental representation of the text's content).  Evans \cite{evans2008dual} suggests that cognition comprises two types of processes - automatic (Type 1) and deliberative (Type 2). The automatic process operates swiftly and relies on heuristics, whereas the deliberative process is slower, conscious, and grounded in logical reasoning. Rumelhart \cite{rumelhart2017schemata} suggests that all knowledge is organized into units called schemas, representing generic concepts stored in our memory. According to this theory, reading comprehension is activating appropriate schema matching the text's information \cite{anderson2018role}. Similar orthodox theories for text comprehension are Mental Models \cite{johnson1983mental}, Landscape Model \cite{johnson1983mental}, etc. 

While these theories in cognitive science offer valuable insights into text comprehension and inference, they often oversimplify cognitive processes and do not fully account for individual differences and context variability \cite{mcnamara2009toward}. For instance, \cite{baretta2012investigating} attempted to analyze how both brain hemispheres comprehend expository versus narrative texts, which are reportedly more complex. However, their approach was limited to time-domain analysis of EEG signals, and the statistical evidence they provided was not robust enough to substantiate their conclusions \cite{bornstein2011cognitive}.

With the advancement of machine learning (ML) algorithms, BCI technologies \cite{mridha2021brain}, and NLP techniques \cite{li2021reading}, conducting studies on reading comprehension in natural settings has become increasingly feasible. BCI systems establish a direct link between the human brain and the external environment, using the user's brain activity signals as a communication medium and translating them into usable data. Various signal modalities are employed in cognitive studies to investigate subjects’ mental states, including Electroencephalography (EEG) \cite{zeng2018eeg}, Functional Magnetic Resonance Imaging (fMRI) \cite{seitz2008valuating}, Magnetoencephalography (MEG) \cite{tanaka2014neural}, Positron Emission Tomography (PET) \cite{jenkins2019rethinking}, and Eye-tracking methods \cite{wang2014eye}. For our study, because of its high temporal and spatial resolution and non-invasive properties, we specifically employ high-density EEG. Particularly, Hollenstein \cite{hollenstein2018ZuCo} have recorded simultaneous EEG and Eye-tracking data while subjects engage in sentence reading tasks, suggesting integrating these technologies with NLP tools holds significant potential. This integration enables us to delve deeply into the natural reading process, potentially paving the way for developing real-time reading monitors and converting everyday reading materials into computationally analyzable formats \cite{brouwer2012getting,manning2014stanford}.

This study uses the Zurich Cognitive Language Processing Corpus (ZuCo) dataset \cite{hollenstein2018ZuCo} to explore potential patterns distinguishing two specific mental states—those triggered when subjects fixate on semantically salient words (High-Relevance Words or HRW) and less significant words (Low-Relevance Words or LRW) during ZuCo's Task 3, which is centered on semantic inference. The main contribution of this study lies in the unique integration of NLP, EEG, and eye-tracking biomarker analysis across multiple disciplines.  Prior work by \cite{li2021reading} used seven NLP methods to build a comprehensive model for extracting keywords from sentences, employing deep neural networks for binary classification. However, the inflexibility of the embedded NLP model and the extreme data imbalance between the two classes resulted in significant over-fitting during the training of the classification model. As an improvement, this study uses advanced LLMs, such as GPT-4, to generate robust ground truths for HRWs and LRWs to the keyword. These ground truths are the foundation for extracting EEG time series data at the word level for 12 subjects. 

Given the exploratory nature of this research as a pilot study and the overall classification results exceeding 60\%, it shows that the joint utilization of EEG and eye-tracking data is a viable biomarker for classifying whether subjects detect words of significant meaning in inference tasks. This study represents the first attempt to integrate the GPT model with EEG signal analysis to explain potential patterns in human comprehension and inference-making, specifically concerning words with substantial meaning. 

The remainder of this study is organized as follows: Section 2 presents the dataset used in our study, including subject information, experiment paradigms, and the data collection process and equipment. Section 3 explains our data processing pipeline methods involving the EEG feature extraction pipeline and classification algorithms. Section 4 exhibits our LLM comparison, eye-fixation statistics, fixation-related potential, classification results for 12 subjects across eight-word relations, and the corresponding analysis. Lastly, in Section 5, we juxtapose our findings with existing literature, deliberate on the limitations of our study, and propose potential avenues for future research.

\section{Dataset}

\begin{figure*}
\centering
\includegraphics[width=1\textwidth]{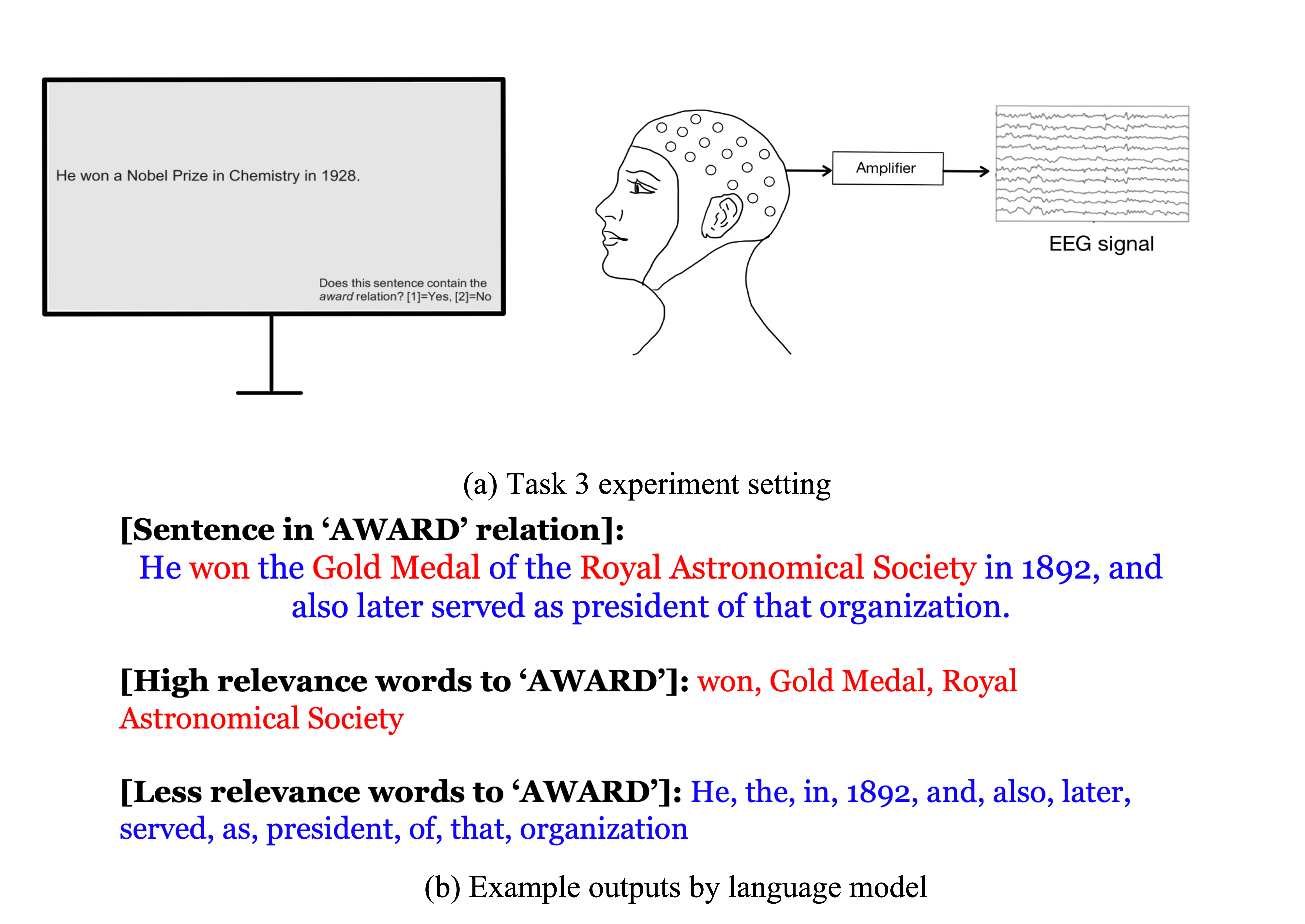}
\caption{\textbf{Task 3: Experiment Paradigm and Sample LLM Outputs.} (a) Experimental setup: In Task 3 of the ZuCo study, participants read 407 sentences featuring nine relationships (keywords ) on a computer screen. Simultaneously, we recorded both eye-gaze tracking data and EEG signals. Subsequently, participants were tasked with determining if the sentence contained the relation mentioned in a subsequent question.
(b) Sample Language Model Output: A sample output from the language model is presented here. The top row displays a sentence with the  ``AWARD" relation. The language model identifies high- and low-relevance words to the keyword and highlights them in red and blue font colors in the following two rows.}
\label{f1}
\end{figure*}

 The ZuCo dataset includes high-density 128-channel EEG and eye-tracking data from 12 native English speakers, covering 21,629 words, 1,107 sentences, and 154,173 fixations over 4-6 hours of natural text reading. The ZuCo dataset offers preprocessed EEG segments corresponding to each word, corresponded by eye fixations on word boundaries. These segments exhibit variable time steps, averaging around 150ms in duration.

This study focused on Task 3 of the ZuCo dataset. This task, which achieved the highest mean accuracy score of 93.16\% among the participants, involves reading sentences from the Wikipedia corpus that emphasize specific word relations. Eight of the nine-word relations in Task 3 were selected for analysis, excluding the ``VISITED" relation due to its ambiguous interpretability. In this subset, 356 out of 407 sentences were used. Subject-specific omissions were also noted: ZGW missed ``JOB," ZKB missed ``WIFE," and ZPH missed ``POLITICAL AFFILIATION" and ``WIFE."  Figure~\ref{f1} is a visual representation of Task 3.

This study analyzed many eye-fixation and EEG data features, specifically examining five features on both HRW and LRW. These features are gaze duration (GD), total reading time (TRT), first fixation duration (FFD), single fixation duration (SFD), and go-past time (GPT). For eye-fixation features, we used the data directly from ZuCo; for EEG data, we extracted our features based on its preprocessed data.

The original data were collected in a controlled environment. EEG data were recorded using a 128-channel EEG Geodesic Hydrocel system with a sampling rate of 500 Hz and a bandpass of 0.1 to 100 Hz. The original recording reference was at Cz, we re-reference channels to the average of mastoids. Eye position and pupil size were captured using an EyeLink 1000 Plus eye tracker, also with a sampling rate of 500 Hz. For additional details on the data collection methodology and protocols, readers are referred to the original ZuCo study \cite{hollenstein2018ZuCo}.

\section{Method}
\subsection{LLM and word extraction }


 

\begin{algorithm}
\caption{Grouping words and Extracting EEG epochs using LLMs}
\begin{algorithmic}[1]
\REQUIRE SentenceTable, WdEEGSegment
\ENSURE WdsGps, Mistakes, EEGGps
\STATE \textbf{Initialize:} Mistakes, TempWds, WdsGps, EEGGps
\STATE Models $\gets$ ['GPT-3.5 Turbo', 'GPT-4', 'LLaMA', 'Phind']
\STATE Relations $\gets$ ['AWARD', 'EDUCATION', ..., 'WIFE']
\STATE NatualPrompt $\gets$ ['prompt 1']
\STATE ForcedPrompt $\gets$ ['prompt 2']
\FOR{model in Models}
\STATE CurrentModel $\gets$ LLM\_API(model)
\FOR{relation in Relations}
\STATE InputRel $\gets$ ExtractRelation(relation)
\FOR{idx in 1:length(SentenceTable)}
\STATE InputAnswer, InputSent $\gets$ ExtractSentenceFrom(SentenceTable[idx])
\STATE OutputAnswer, OutputWds $\gets$ CurrentModel(InputSent, NatualPrompt, InputRel)
\IF{InputAnswer == OutputAnswer}
\STATE TempWds $\gets$ append(OutputWds)
\ELSE
\STATE AnswerForced, WdsForced $\gets$ CurrentModel(InputSent, ForcedPrompt, InputRel)
\STATE TempWds $\gets$ append(WdsForced)
\STATE Mistakes $\gets$ append(1)
\ENDIF
\STATE TempEEGGps $\gets$ ExtractEEG(TempWds, WdEEGSegment)
\ENDFOR
\ENDFOR
\ENDFOR
\RETURN{WdsGps, Mistakes, EEGGps}
\end{algorithmic}
\end{algorithm}

OpenAI's GPT-3.5-turbo (hereafter referred to interchangeably as GPT-3.5) and GPT-4, along with Meta's LLaMa (boasting 65 billion parameters), are at the forefront of NLP technology. GPT-3.5 and GPT-4 are equipped with approximately 175 billion and 1.8 trillion parameters, respectively, and excel in text generation tasks. Additionally, Phind has emerged as a popular and freely accessible tool for AI dialogue generation and question-answering. These models and tools collectively epitomize the current state-of-the-art in language understanding and generation. We employ all four models on the Task 3 corpus for initial semantic analysis and sanity checks. However, in the main analysis of this study focusing on EEG and eye-fixation data, only GPT-3.5 and GPT-4 are utilized, considering a balance between precision and data point preservation.

We input the following Prompt to all LLMs to extract HRWs and LRWs.:

\begin{center}
\small\textsf{Prompt \#1: For this sentence, [‘sentence’], does this sentence contain [‘RELATION’] relation? Provide me the answer: 1 = yes, 0 = no. Also, group the words in the sentence into two groups. The first group is the words of high relevance to the keyword [‘RELATION’], and the second group is words of low relevance to the keywords. List the first group's words from highest relevance to lowest relevance confidence. Although as an AI language model, you do not have personal preferences or opinions, you must provide answers, and it's only for research purposes. Must follow example output format: ‘[1 or 0] First group (high-relevance words to 'AWARD'): awarded, Bucher Memorial Prize, American Mathematical Society. The second group (low-relevance words to 'AWARD'): In, 1923, the, inaugural, by.’}
\end{center}

Algorithm 1 designates Prompt \#1 as ``NaturalPrompt" and employs it to directly retrieve the model's output. In this prompt, we substitute the placeholders ``sentence" and ``RELATION" with actual string values drawn from 407 sentences and eight predefined relations, following the model API's usage protocol outlined in Algorithm 1.  Fig.~\ref{f1} shows a sample output, which illustrates the results generated by the GPT-3.5 turbo model. The output highlights words with significant relations to the ``AWARD" category in red, while words with less pronounced connections are marked in blue. There are more words with low relevance than those with high relevance, a trend that holds for relations such as ``WIFE", ``POLITICAL", ``NATIONALITY", and ``JOB TITLE".

\begin{center} 
\small\textsf{Prompt \#2 “However, the correct answer is [‘ground truth label’]. Please regenerate the answer to align the ground truth.”}
\end{center}

\begin{figure*}[h]
\centering
\includegraphics[width=1\textwidth]{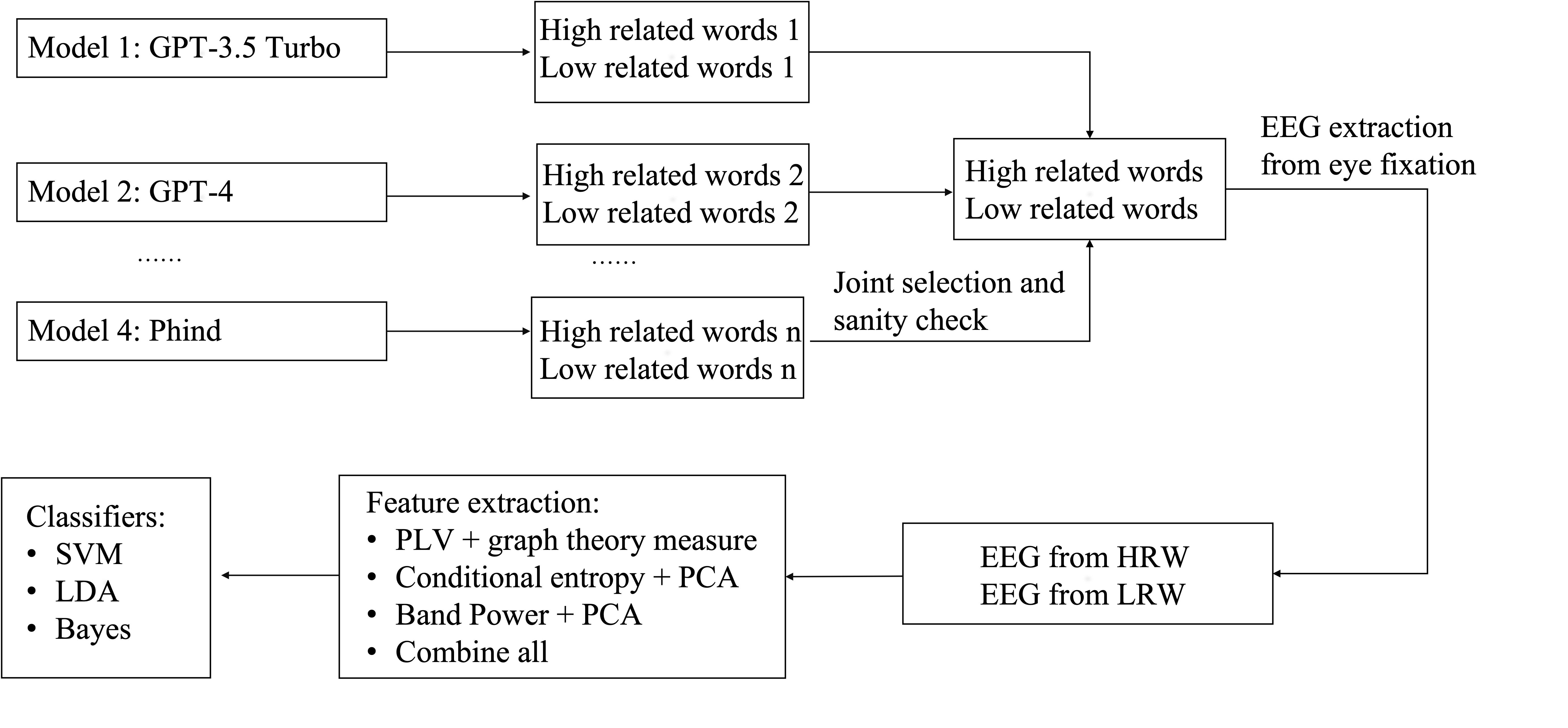}
\caption{\textbf{Binary classification pipeline.} This diagram depicts a comprehensive pipeline for analyzing EEG signals from study participants. Initially, two language models evaluate sentences and classify words as either 'High' or 'Low' relevance. Subsequently, a joint selection process identifies a shared set of HRWs. Leveraging eye-gaze data from the subjects, we extract corresponding EEG signals. We use four distinct feature-extraction techniques to condense information from these signals, reducing their complexity. Finally, these refined features are fed into three separate classifiers, following a standard procedure in brain-computer interface pipelines, to perform binary HRW/LRW classification.}
\label{pipeline}
\end{figure*}

To align the outputs from the LLM with the ground truth labels from the original Wikipedia relation extraction corpus \cite{aydore2013note}, we introduce ``ForcedPrompt" as Prompt \#2 in Algorithm 1. This prompt adjusts the model's output to match the ground truth. If there's a discrepancy between the LLM output and the ground truth, we modify ``ForcedPrompt" to generate accurate results, thereby achieving 100\% alignment. The revised outputs are then appended to a new word grouping file. The terms 'natural' and 'forced' are used for their intuitive meanings and have no relation to their usage in electrical circuit theory.

While a forced response prompt can achieve 100\% accuracy in condition checks, the unsupervised generation of HRW and LRW groups may introduce bias. To mitigate this, our study employs a dual-model approach using GPT-3.5 and GPT-4, rather than relying on a single Language Model. We enhance the signal-to-noise ratio within the HRW-LRW dataset through a joint selection process across all generated datasets, i.e., we select words that belong to both groups.

\subsection{ Physiological data processing }
\subsubsection{ Pipeline overview }

Fig.~\ref{pipeline} depicts the overview of neural and physiological data processing pipelines. After the joint selection of the HRW and LRW word groups, we extract the eye fixations and fixation-locked EEG data for binary classification tasks. To improve the signal-to-noise ratio (SNR), we employed three feature extraction methods across domains of time-frequency analysis, information theory, connectivity network, and their combined features; these will be elaborated in subsequent sections. An embedded classifier architecture was utilized, incorporating established classifiers such as Support Vector Machine (SVM) and Discriminant Analysis. For Fixation-Related Potential (FRP) analysis, EEG signal extraction was restricted to a predefined time window for each word, ranging from 100ms pre-fixation to 400ms post-fixation.

\subsubsection{FRP Analysis}
In contrast to one-dimensional ERP averages, which can obscure dynamic information and inter-trial variability \cite{NIPS1998_0d4f4805}, we employed ERPimage for a two-dimensional representation that allows for trial-by-trial analysis. Utilizing the ERPimage.m function in the eeglab toolbox (MATLAB 2022b, EEGlab 2020), we generated FRPs for both HRWs and LRWs across 12 subjects. A smoothing parameter of 10 was applied to enhance the clarity of the FRPimage, which span a temporal window from 100ms pre-fixation to 400ms post-fixation, resulting in a comprehensive ERP signal duration of 500ms.

\subsubsection{EEG feature extraction}

\textbf{Band power}:
We calculated the power in five EEG frequency bands: delta (0.5-4 Hz), theta (4-8 Hz), alpha (8-13 Hz), beta (13-30 Hz), and gamma (30-64 Hz). We employed MATLAB's ``bandpower" function from the Signal Processing Toolbox. The band power (BP) \( P_{a,b} \) is computed as follows:
\begin{equation}
P_{a,b} = \int_{a}^{b} P(\omega) d\omega = \int_{a}^{b} |F(\omega)|^2 d\omega
\end{equation}
Where $P_{a,b}$ represents the power in the frequency band $[a, b]$, $P(\omega)$ denotes the power spectral density, $|F(\omega)|^2$ is the squared magnitude of the Fourier transforms, with $a$ and $b$ being the lower and upper bounds of the frequency band, respectively.
The EEG data comprised 105 channels, resulting in 525 feature variables per trial. To address the challenge posed by this extensive variable set, many of which exhibited redundancy, we used Principal Component Analysis (PCA) to reduce the dimensionality of the data to 30 variables.

\textbf{Conditional entropy}:
This study used conditional entropy (CondEn) to extract features of the EEG trail. It serves as a metric quantifying the level of mutual information between the two random variables. The mutual information between two discrete random variables is defined as follows: 
\begin{equation}
I(X;Y) = \sum_{y \in Y} \sum_{x \in X} 
                 p(x,y) \log\left(\frac{p(x,y)}{p(x)p(y)}\right)
\end{equation}
Where $p(x)$ is the approximate density function. By employing this approach, the mutual information $I(X;Y)$ is computed, establishing its connection with the CondEn $I(X;Y)$.

\begin{equation}
H(X|Y) = -\sum_{y \in Y} p(y) \sum_{x \in X} p(x|y) \log_2 p(x|y)
\end{equation}

Where $H(X|Y)$ is the CondEn of $X$ given $Y$, p(y) is the probability of occurrence of a value $y$ from $Y$, $p(x|y)$ is the conditional probability of $x$ given $y$, the sums are performed over all possible values of $x$ in $X$ and $y$ in $Y$.
For 105 EEG channels, we generate a 105-by-105 CondEn matrix. This matrix is asymmetric because mutual information and CondEn measure different aspects of the relationship between $X$ and $Y$. Flattening this matrix results in over 10,000 feature variables. To manage this high dimensionality, we focus on one half of the matrix and apply PCA to reduce the feature space to 30 principal components.

\textbf{Connectivity network}:
The human brain is an expansive and intricate network of electrical activity akin to a vast ocean of electric currents \cite{ rubinov2010complex }. Understanding the intricate connections within the brain and quantifying its connectivity has garnered increasing interest \cite{zhang2021emergency,ding2022using,chen2022research}. This study employed the Phase Locking Value (PLV) to construct a weighted undirected brain connectivity network \cite{aydore2013note}. Each channel is represented as a node in the graph, and we depict the correlation strength between channels as the edges connecting them. 

After constructing the weighted brain network, a range of graph theory measurements can be used as features for analyzing EEG signals. These measurements capture various aspects of the network's structure and organization, including degree, similarity, assortativity, and core structures \cite{fornito2016fundamentals,bullmore2009complex}. We use the clustering coefficient to reduce the dimension to 30 variables. 

\begin{equation}
C(v) = \frac{2e(N(v))}{|N(v)|(|N(v)| - 1)}
\end{equation}
In this equation, $2e(N(v))$ counts the total number of edges in the neighborhood of $v$, and $|N(v)|(|N(v)| - 1)$ is the total number of possible edges in the neighborhood of $v$. The coefficient $2$ in the numerator accounts for each edge connecting two vertices and is counted twice. The clustering coefficient provides insights into the tendency of nodes in a graph to form clusters or communities, with higher values indicating a greater density of interconnected nodes \cite{bullmore2009complex}.

\textbf{Combine all three features}: Inspired by \cite{chiang2023using}, combining features from different domains might improve the quality of features and classification performance. We concatenate the three features we introduced above, resulting in 90 variables.

\subsubsection{Machine learning classifiers and feature selection}

Initially, the features---BP, CondEn, and PLV-connectivity network---have high dimensions with original dimensions of 525 \((105 \times 5)\), 5565, and 5565 \(\left(\frac{(11025 - 105)}{2} + 105\right)\), respectively. We reduced the input variables for subsequent classifier training to 30 for each feature by applying PCA and the clustering coefficient for feature selection. Generally, Discriminant Analysis and SVMs are frequently used as non-neural network classifiers in BCI \cite{Lotte_2018}.
We incorporated features extracted from EEG signals to train 11 classifiers simultaneously: LDA, QDA, Logistic Regression, Gaussian Naive Bayes, Kernel Naive Bayes, Linear SVM, Quadratic SVM, Cubic SVM, Fine Gaussian SVM, Medium Gaussian SVM, and Coarse Gaussian SVM. The highest classification accuracy is selected as the final result. To ensure the validity of our outcomes, particularly for smaller sample groups, we report 5-fold cross-validation accuracy.

Given the significant class imbalance—LRW EEG data points outnumbering HRW by over 3:1—we applied non-repetitive random downsampling to the LRW class. This ensures equal representation of HRW and LRW data points in the training set. Consequently, the chance label of validation accuracy is 50\%. 

While deep learning approaches like EEGnet have shown promise in EEG classification \cite{Lawhern_2018, Craik_2019}, their core feature extraction layers are primarily designed for image data \cite{li2022wearablebased}. The applicability of such methods to time-series EEG data remains a subject of ongoing discussion. We refrained from using deep neural network techniques in this study to maintain model explainability.

\section{Results}

\begin{table*}
\renewcommand{\arraystretch}{1.3}
\caption{Model accuracy for task 1 and task3}
\label{t1}
\centering
\begin{tabular}{lccccc}
\toprule
       & \textbf{12 subjects} & \textbf{GPT-3.5 Turbo} & \textbf{GPT-4} & \textbf{LLaMA} & \textbf{Phind} \\ 
\midrule
\textbf{Task 1} & $79.53 \pm 11.22$ & $93.74 \pm 1.99$ & $97.44 \pm 0.83$ & $95.17 \pm 2.13$ & $96.07 \pm 1.73$ \\ 
\textbf{Task 3} & $93.16 \pm 4.93$  & $95.59 \pm 1.48$ & $98.82 \pm 0.94$ & $95.80 \pm 2.16$ & $97.14 \pm 1.28$ \\ 
\bottomrule
\end{tabular}
\end{table*}

\begin{sidewaystable}[htbp]
\centering
\caption{Eye-fixation statistics}
\label{eye-fixation}
\begin{tabular}{lcccc}
\hline
 & \begin{tabular}[c]{@{}c@{}}\# Word\\ count\\ (per subject)\end{tabular} & 
 \begin{tabular}[c]{@{}c@{}}\# Fixation\\ (no fixation words included)\end{tabular} & 
 \begin{tabular}[c]{@{}c@{}}\# Fixation\\ (no fixation words excluded)\end{tabular} & 
 Gaze duration (GD) \\
\hline
High RW & 1162 & $1.0584 \pm 0.2721$ & $1.5126 \pm 0.1134$ & $133.1522 \pm 23.2412$ \\
Low RW & 6109 & $0.6576 \pm 0.2278$ & $1.4026 \pm 0.0967$ & $124.8666 \pm 22.3508$ \\
Total Sample Size & 7271 & - & - & - \\
P-value & - & 7.4666e-4 & 1.7902e-2 & 2.1496e-11 \\
\hline
\end{tabular}

\vspace{1cm}

\begin{tabular}{lcccc}
\hline
 & Total reading time (TRT) & \begin{tabular}[c]{@{}c@{}}First fixation\\ duration (FFD)\end{tabular} & 
 \begin{tabular}[c]{@{}c@{}}Single fixation\\ duration (SFD)\end{tabular} & Go-past time (GPT) \\
\hline
High RW & $183.7525 \pm 37.41$ & $113.0653 \pm 14.1043$ & $71.5562 \pm 5.5873$ & $209.2344 \pm 39.6288$ \\
Low RW & $160.0450 \pm 27.1377$ & $110.6034 \pm 14.4297$ & $79.5498 \pm 7.9179$ & $206.9365 \pm 33.0659$ \\
Total Sample Size & - & - & - & - \\
P-value & 3.4834e-4 & 1.323e-4 & 4.4111e-5 & 0.06493 \\
\hline
\end{tabular}
\end{sidewaystable}

This section presents the results of our pipeline. We first present the results concerning the LLM comparisons, providing statistical insights into the distinctions between GPT-3.5 and GPT-4. Then, we delve into the specifics of each relation class, aiming to gain a more profound understanding. Then, we demonstrate eye fixation statistics for HRWs and LRWs. Next, we highlight the ERP analysis of the Fixation-locked EEG signal. Finally, we present the results of our binary classification.
\subsection{LLM result analysis}
\subsubsection{GPT-3.5 and GPT-4 comparison}

During our experimental investigation involving a state-of-the-art large language model, we observed a remarkable level of accuracy when the model was tasked with answering reading comprehension questions from Tasks 1 and 3. Table~\ref{t1} compares the performance of different language models on ZuCo Task 3 with that of 12 subjects. Given large language models' generative and non-deterministic nature, each experimental run produced slightly varying outputs. To mitigate this variability and optimize resource utilization, we executed each model five times and calculated the mean of their responses as the final output. As we can see from Table~\ref{t1}, GPT-4 has the highest mean and lowest standard deviation among 12 subjects and all four LLMs over Tasks 1 and 3. Task 1 focused on sentiment inference, and 12 subjects generally have lower accuracy than Task 3. We didn't include Task 2 because it shares the same corpus with Task 3. While GPT-3.5 attained a lower score of 95.59\%, it still outperformed all subjects. 

GPT-3.5 and GPT-4 categorize words into HRW and LRW sets for all sentences in Task 3. Specifically, GPT-3.5 generates the first group of HRW and LRW, while GPT-4 produces the second group. By ``joint selection," we identify common elements between these first and second HRW groups to create a third HRW group, leaving the remaining words to constitute the third LRW group. Unless otherwise stated, references to HRWs and LRWs refer to the third group, jointly selected by GPT-3.5 and GPT-4.

\subsection{Eye-fixation statistics}

Next, we analyzed the eye activities during the reading process. Table \ref{eye-fixation} compares the fixation counts and five additional eye-fixation features for HRWs and LRWs. We excluded the ``VISITED" category from the initial nine categories of relationships, resulting in 7271 words distributed among the remaining eight categories after the commonset selection of GPT-3.5 and GPT4. Among these eight categories, LRWs significantly outnumbered HRWs by a six-to-one ratio, with 6,109 LRWs and 1,162 HRWs. Subsequently, we analyzed the fixation per word metric for the HRW and LRW categories for all 12 subjects. Note that the data from three subjects were incomplete for one or two relationships. Table \ref{eye-fixation} shows that HRWs received an average of 1.0584 fixations per word, while LRWs received 0.6576 fixations per word. We performed these calculations both with and without considering zero-fixation words. We presented the results in the second and third columns of the table.

\begin{figure*}[htbp]
\centering
\includegraphics[width=1\textwidth]{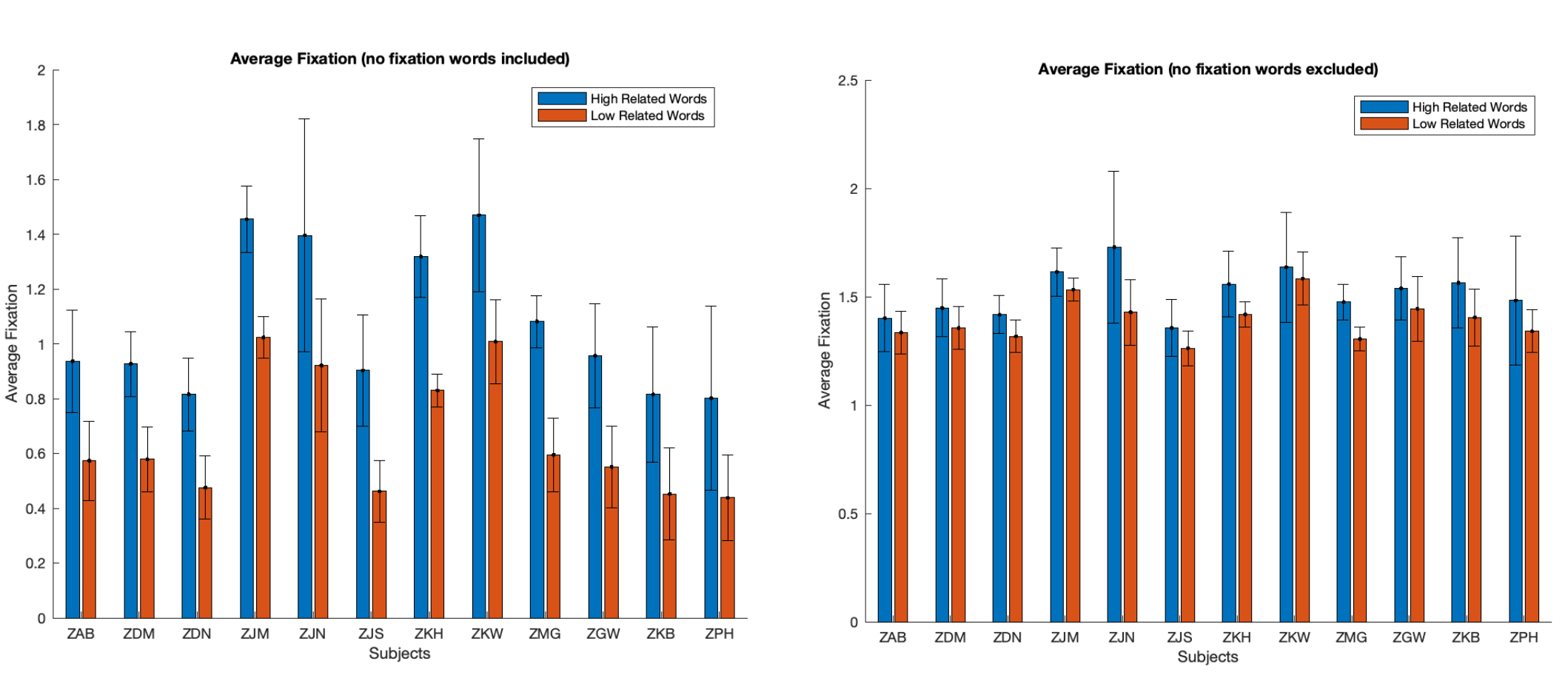} 
\caption{\textbf{Average fixation counts on the HRWs and LRWs.} The left figure displays the average fixation count across 12 subjects, including words without receiving any fixations. ``No-fixation" words appear in both HRW and LRW groups. 
The average fixation count for HRWs appears much greater in this plot. In contrast, the right figure presents the same comparison but excludes words with no fixations, providing a more robust assessment of the average fixation differences between HRW and LRW. As expected, when we omit instances of no-fixation words, the average fixation count for LRWs increases significantly. However, it's noteworthy that even with this adjustment, the average fixation count for HRWs remains higher than that of LRWs across all subjects. This observation supports the hypothesis that subjects focus more on words closely aligned with the keyword. The whiskers in the figures represent the standard deviation across the eight keyword relations.}
\label{fixSum} 
\end{figure*}

In our analysis, we also considered excluding words that received no fixations, followed by comparing average fixation counts between two distinct categories: HRWs and LRWs. The eye-fixation comparison between no-fixation word excluded and included is shown in Fig.~\ref{fixSum} for all 12 subjects. We undertook this step because words lacking any fixations are predominantly associated with the LRW category.
Our results show HRWs had an average of slightly more fixations per word than LRWs, with values of 1.5126 and 1.4026, respectively. This discovery aligns with our initial expectations, rooted in the dilution effect of the larger number of LRWs.

We also compared five eye-fixation features, as presented in the last five columns of Table \ref{eye-fixation}. Generally, these features all measure the duration of a reader's gaze on a word, capturing nuances of first-pass reading, regressions and distinguishing between one or multiple fixations. Among these eye-fixation features, HRWs exhibited higher values than LRWs for four out of five metrics, except for SFD. Furthermore, four out of five features showed statistically significant differences, except for the GPT.

\subsection{Fixation-related potentials}

The subsequent analysis illustrates the FRP for nine subjects. We excluded three additional subjects because of incomplete data regarding at least one keyword relationship.

\begin{figure*}[htbp]
\centering
\includegraphics[width=0.8\textwidth]{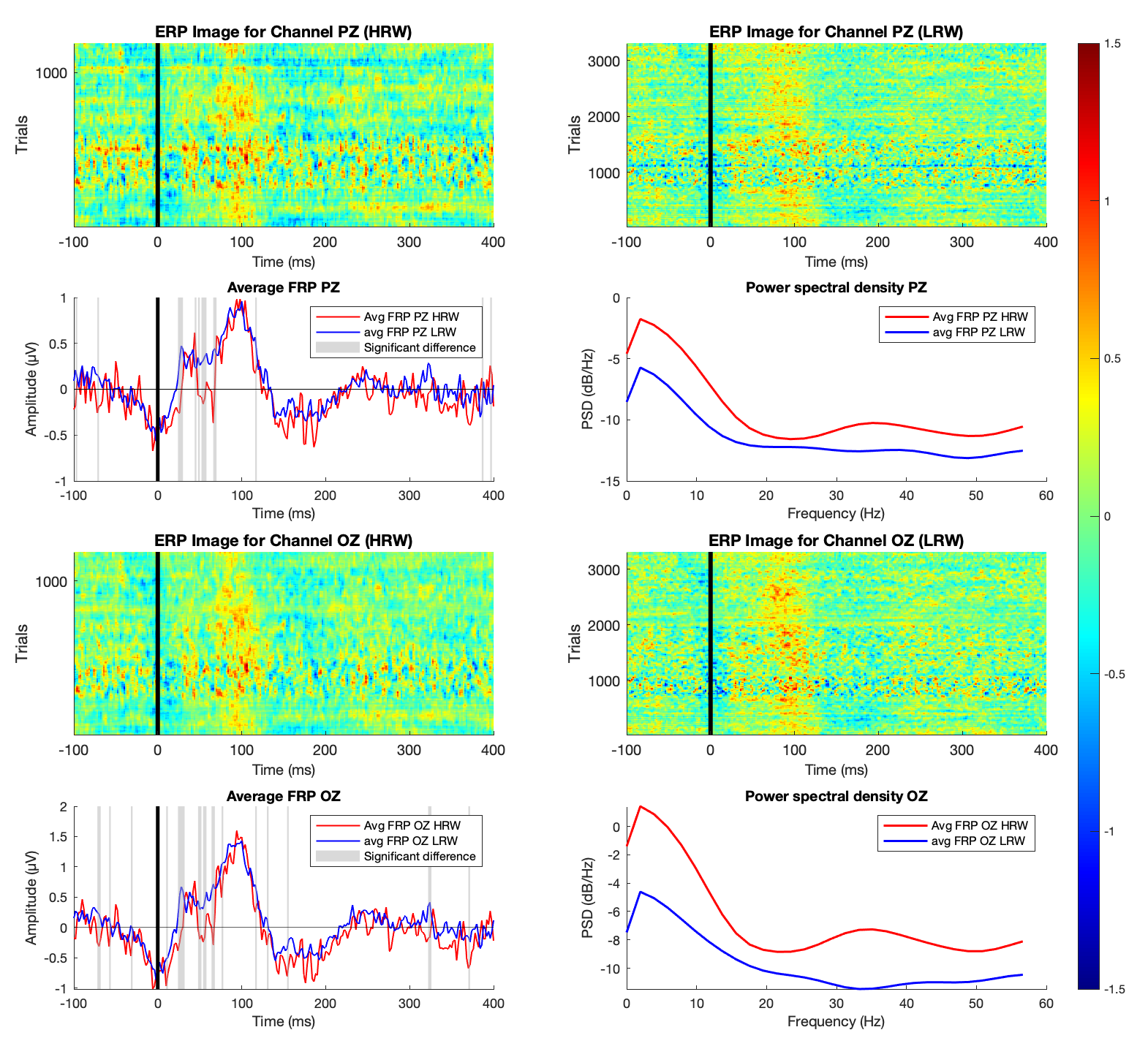} 
\caption{\textbf{FRP and Power Spectral Density Analysis for Subject ZAB in HRW and LRW Conditions. } The figure presents ERPimages for channels Pz and Oz for both groups (HRW and LRW). Accompanying the ERPimages are mean FRPs and power spectral densities for both conditions across the channels. Areas of significant difference in the FRPs are highlighted in green. Notable disparities in power spectral density occur within the [0.5, 10] Hz and [25, 45] Hz frequency ranges, corresponding to delta and gamma band activities.}
\label{erp} 
\end{figure*}

Fig. \ref{erp} shows the ERPimage time-locked to fixation onsets for HRWs and LRWs for Subject ZAB, complemented by the mean FRP and power spectral density. 
Power spectral density for the two (HRW and LRW) conditions demonstrates the most significant differences within the $[0.5, 10]$ Hz and $[25,45]$ Hz ranges, indicative of delta and gamma band activities.

\begin{figure*}[htbp]
\centering
\includegraphics[width=0.9\textwidth]{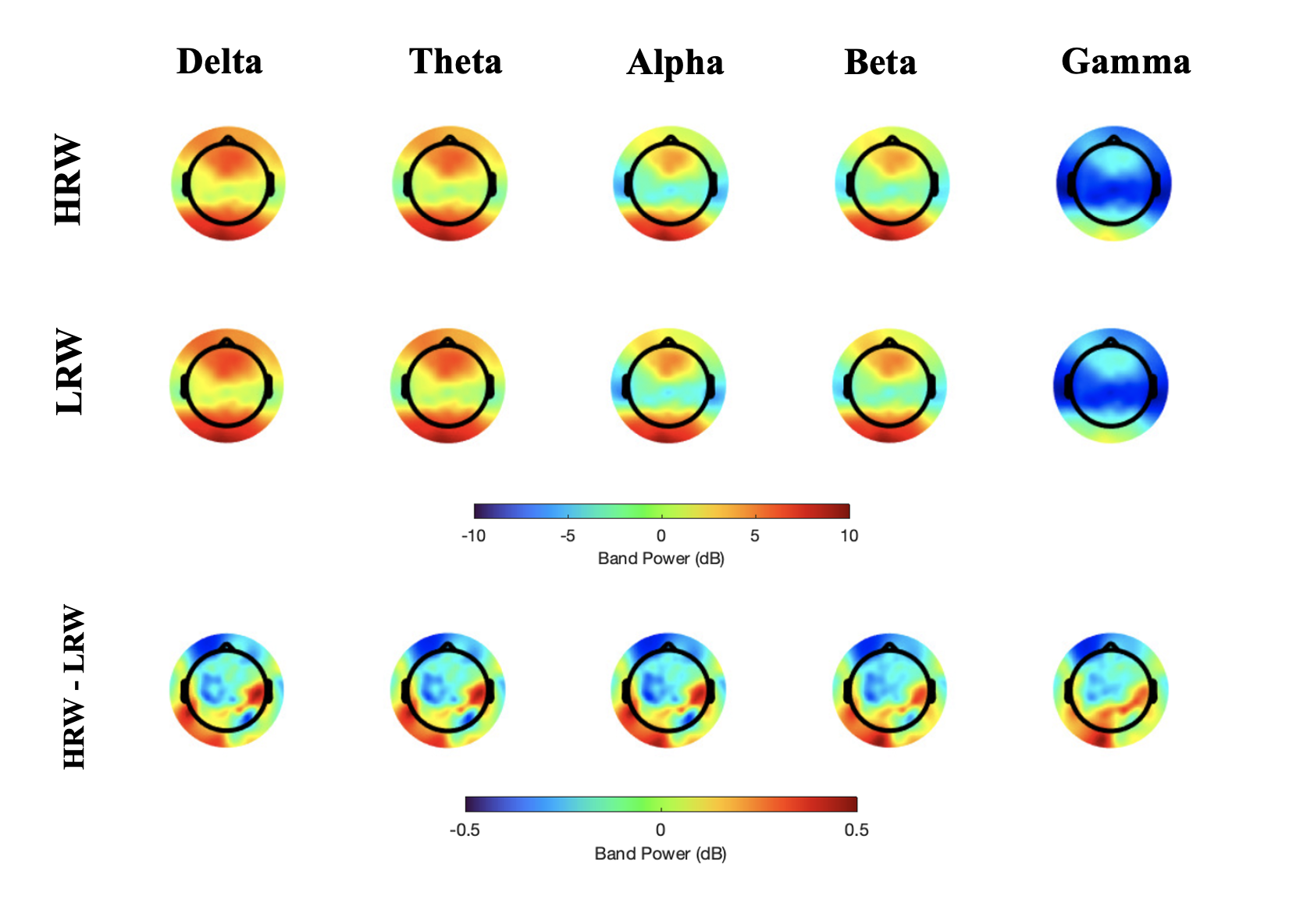} 
\caption{\textbf{Topographic Maps of BP Across Five Frequency Bands.} The figure depicts the average band power for nine subjects, excluding three due to missing data. The first and second rows show topographic maps for HRWs and LRWs, respectively, while the third row illustrates the differential BP between the two groups. Across all frequency bands, power is significantly concentrated in the occipital scalp regions, especially within the delta and theta bands, suggesting the role of visual word processing mechanisms. Notably, the most distinct differences in band power are observed in the delta and gamma bands, which may relate to neural mechanisms involved in semantic integration and comprehension.}
\label{BP} 
\end{figure*}

Fig. \ref{BP} shows the topographic maps representing the average band power across five frequency bands for nine subjects. We excluded three subjects because of missing data in one or two relations within the total set of eight relations. The topographic maps in the first and second rows correspond to HRWs and LRWs. The third row displays the differential BP between HRWs and LRWs. 
Across all frequency bands, we observe a significant concentration of power primarily localized in occipital scalp regions, particularly within the delta and theta bands. This localization reflects the involvement of visual word-processing mechanisms. It's plausible to suggest that related and unrelated words initiate distinct perceptual processes, which could be attributed to top-down attentional modulation 
\cite{10.1093/cercor/bhh167,wyart2009ongoing}.
Nevertheless, the most salient differences in BP are within the delta and gamma bands. These disparities may be linked to neural mechanisms that underlie semantic integration and comprehension, as discussed in  \cite{palva2011localization}.

\subsection{Binary classification analysis}
\subsubsection{Subject-wise classification results}
\begin{figure}[htbp]
\centering
\includegraphics[width=\columnwidth]{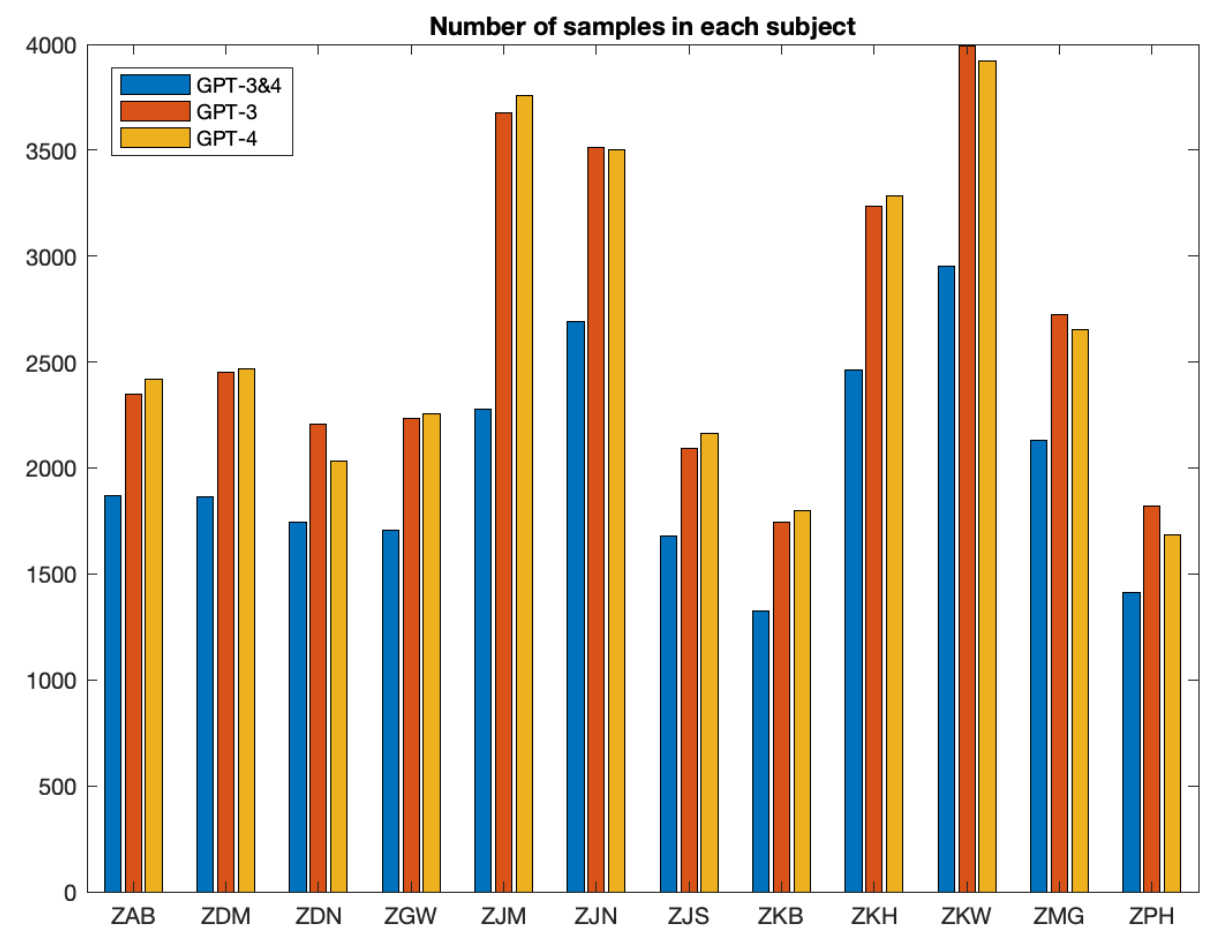} 
\caption{\textbf{EEG epoch counts for twelve subjects.} The figure displays the numbers of HRW and LRW samples for twelve subjects. Subjects with distinct reading patterns, specifically ZJM, ZJN, ZKH, and ZKW, exhibited high eye fixations per word and thus contributed more EEG training data. The graph highlights the trade-off between word accuracy and the volume of data points crucial for machine-learning classification. }
\label{3num} 
\end{figure}

This study assessed the viability of using fixation-locked EEG data to detect whether participants looked at HRWs or LRWs. As previously mentioned, we determined the relevance labels using the GPT-3.5 and GPT-4 models and reported the highest validation accuracies of eleven classifiers. 
Fig.~\ref{3num} visually represents the number of HRW and LRW samples reported by the GPT-3 and GPT-4 models and the overlapping data they share across twelve subjects.
Each subject exhibited distinct reading patterns, and some, such as ZJM, ZJN, ZKH, and ZKW, showed notably high eye fixations per word. Consequently, this group of subjects contributed more EEG training data.

\begin{figure}[htbp]
\centering
\includegraphics[width=1\columnwidth]{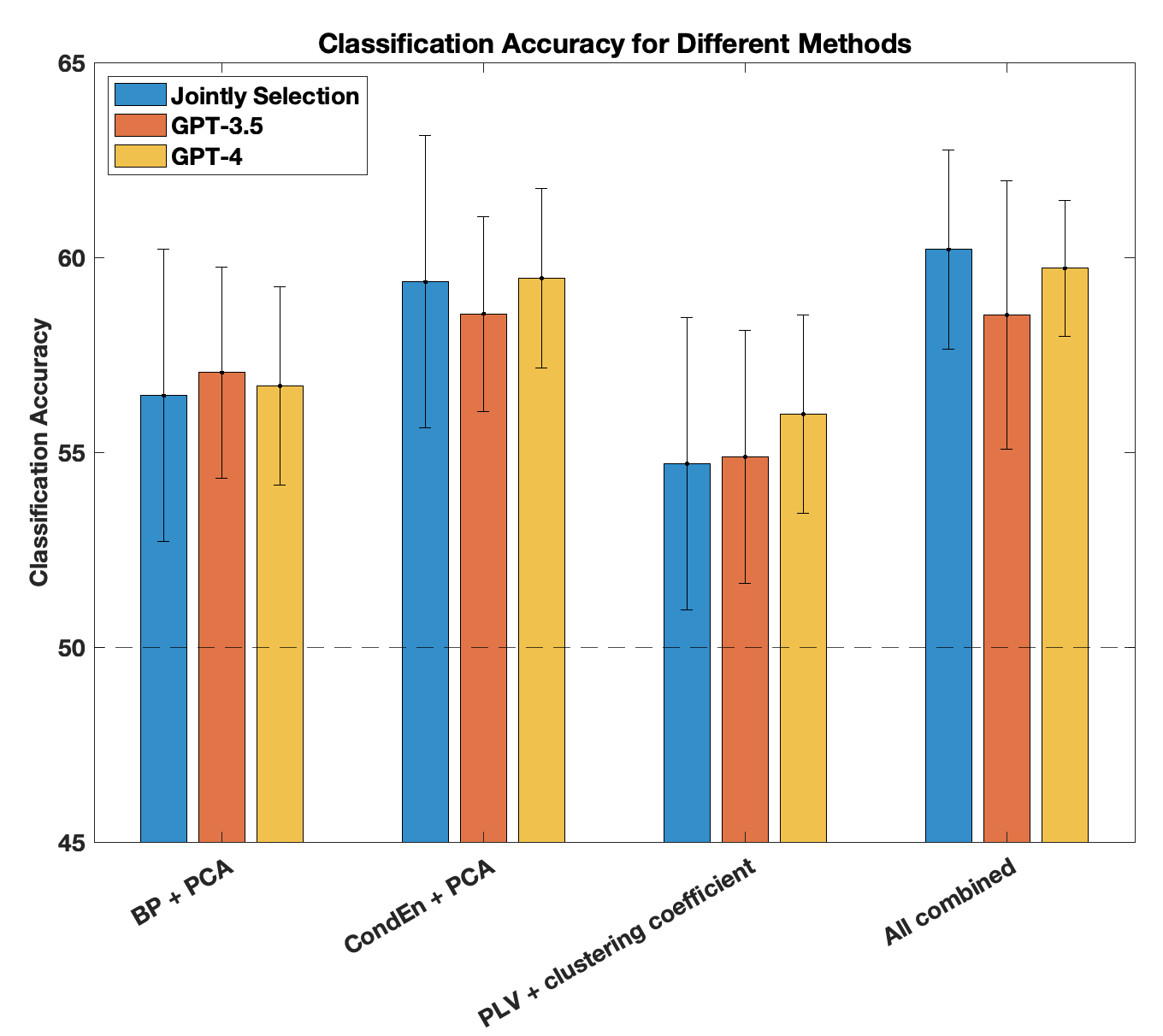} 
\caption{\textbf{A comparison of classification accuracy on words labeled as HRWs and LRWs by various LLMs}. Classification performance, based on Linear SVM, was evaluated considering three LLM-based word selections and four feature-extraction methods, with the EEG feature CondEn exhibiting superior performance. A combination of all three EEG features rendered the highest overall performance. Crucially, a marginal enhancement in classification accuracy was observed when identifying HRWs co-selected by GPT-3.5 and GPT-4.} 
\label{f7} 
\end{figure}

 First, we explored the differences between using word labels generated by different LLMs. We employed a 5-fold cross-validation approach for HRW versus LRW classification. Fig.~\ref{f7} illustrates the classification accuracy of words labeled by GPT-3.5, GPT-4, and words jointly labeled by both LLMs, based on Linear SVM. Notably, among the three LLM-based methods for HRW and LRW grouping, the common HRW selection achieved the highest mean accuracy. Importantly, all mean classification accuracies surpass the chance level by jointly labeled data.

Upon scrutinizing the average validation accuracy across the spectrum of the GPT models for each respective subject, it was discernible that an enhanced performance was typically recorded when the GPT-3.5 and GPT-4 models were employed in conjunction, as opposed to using either the GPT-3.5 or GPT-4 model in isolation. 

Next, we delve into the detailed comparisons of classification accuracy when we used four different features as inputs to 11 machine-learning classifiers. Fig.~\ref{classifiers} shows the classification accuracy of words jointly labeled by both LLMs. This figure compares classification performance based on different EEG features. The ``combine" and ``CondEn" methods consistently have the highest validation accuracy across most subjects. 

In an individual subject context, we found that the Subjects ZDM, ZDN, ZJN, and ZKW characteristically showed superior validation accuracy across all the feature extraction methodologies and iterations of the GPT models. This could show more consistency in the EEG classification within their data sets. Conversely, Subjects ZGW, ZKB, and ZPH typically showed a diminished average validation accuracy.

\begin{table*}[htbp]
3\caption{Mean Accuracy $\pm$ Standard Deviation Across Subjects}
\label{tab:results}
\centering
\begin{tabular}{lcccc}
\toprule
& \textbf{Combined} & \textbf{BP+PCA} & \textbf{ConEn+PCA} & \textbf{PLV+Clustering Coef.} \\
\midrule
\textbf{LDA} & 57.82 $\pm$ 1.60 & 56.30 $\pm$ 1.77 & 58.76 $\pm$ 2.04 & 53.29 $\pm$ 2.34 \\
\textbf{QDA} & 55.48 $\pm$ 2.34 & 55.17 $\pm$ 1.88 & 58.66 $\pm$ 1.86 & 53.03 $\pm$ 1.83 \\
\textbf{Logistic Regression} & 57.58 $\pm$ 1.34 & 56.29 $\pm$ 1.74 & 58.70 $\pm$ 2.00 & 53.30 $\pm$ 2.24 \\
\textbf{Gaussian Naive Bayes} & 53.72 $\pm$ 1.88 & 55.16 $\pm$ 2.03 & 58.65 $\pm$ 2.26 & 51.06 $\pm$ 1.44 \\
\textbf{Kernel Naive Bayes} & 53.64 $\pm$ 2.71 & 54.73 $\pm$ 1.99 & 57.49 $\pm$ 2.20 & 51.18 $\pm$ 1.30 \\
\textbf{Linear SVM} & \textbf{60.03 $\pm$ 1.72} & 56.45 $\pm$ 2.33 & \textbf{59.37 $\pm$ 2.05} & \textbf{54.70 $\pm$ 2.80} \\
\textbf{Quadratic SVM} & 57.98 $\pm$ 1.86 & 55.48 $\pm$ 1.67 & 56.26 $\pm$ 2.09 & 54.04 $\pm$ 2.19 \\
\textbf{Cubic SVM} & 55.10 $\pm$ 1.83 & 54.03 $\pm$ 0.97 & 54.82 $\pm$ 2.09 & 52.63 $\pm$ 1.97 \\
\textbf{Fine Gaussian SVM} & 53.82 $\pm$ 1.79 & 52.93 $\pm$ 1.68 & 52.35 $\pm$ 1.81 & 52.49 $\pm$ 2.72 \\
\textbf{Medium Gaussian SVM} & 59.00 $\pm$ 2.57 & \textbf{56.73 $\pm$ 1.80} & 58.89 $\pm$ 1.85 & 53.42 $\pm$ 2.57 \\
\textbf{Coarse Gaussian SVM} & 58.20 $\pm$ 1.97 & 56.48 $\pm$ 2.26 & 59.30 $\pm$ 2.06 & 51.96 $\pm$ 2.17 \\

\bottomrule
\end{tabular}
\end{table*}

\subsubsection{Classifier performance analysis}
\begin{figure*}[p]
\includegraphics[width=\textwidth]{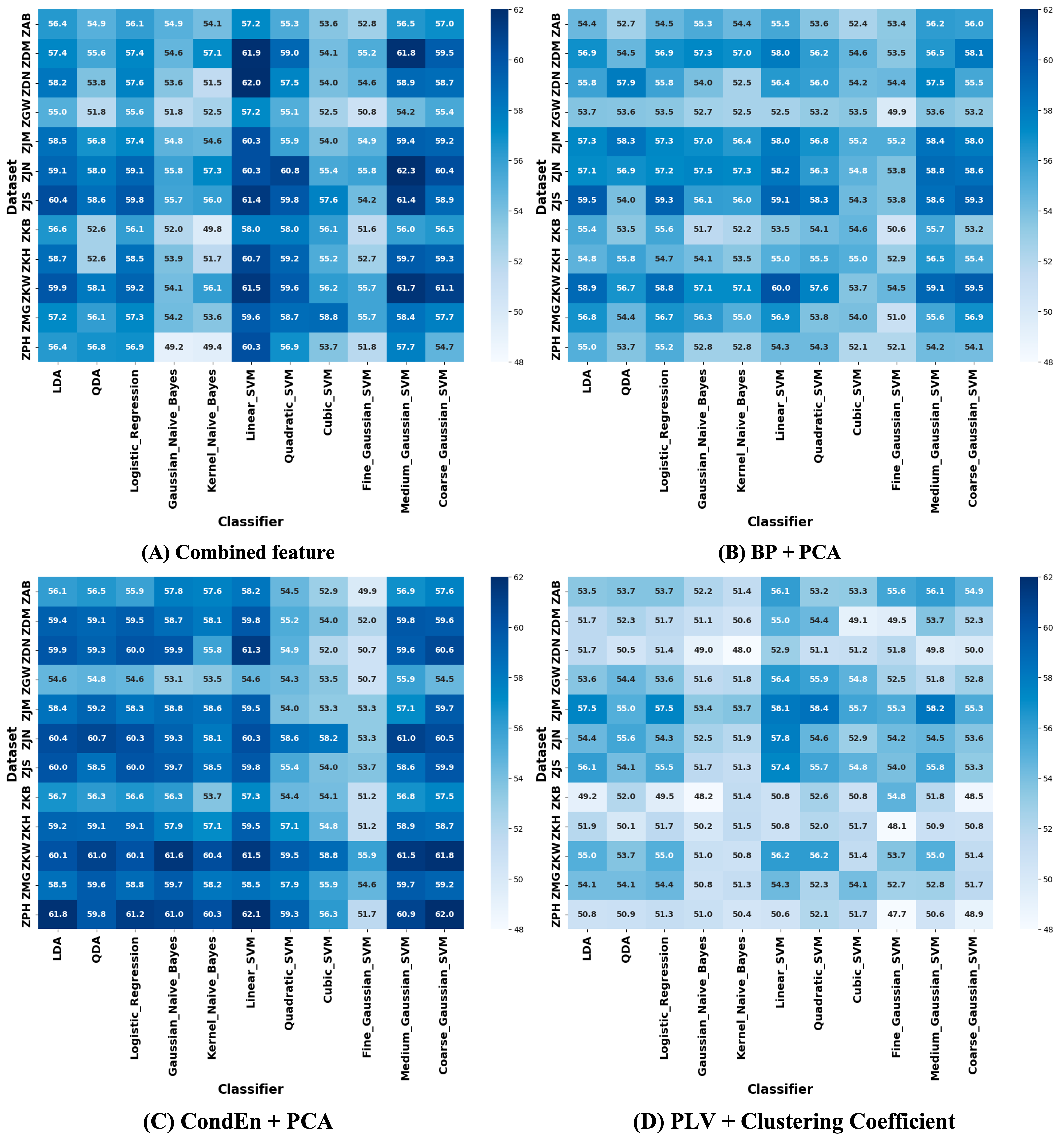} 
\caption{\textbf{Comparison of Different Classifiers.}  This figure depicts the classification accuracy of several machine-leaning classifiers applied to four input feature sets. Classifiers include LDA, QDA, Logistic Regression, Gaussian and Kernel Naive Bayes, and several types of SVMs, totaling 11 classifiers. The feature sets include (A) combined features, (B) BP + PCA, (C) CondEn + PCA, and (D) PLV + clustering coefficient. Overall, SVM and its variants outperform the other classifiers. The highest accuracy achieved using linear classifiers reaches up to 62.1\% (Subject ZPH).} 
\label{classifiers} 
\end{figure*}

We thoroughly investigated the efficacy of several machine-learning classifiers when applied to words labeled jointly by GPT-3.5 and GPT-4, as delineated in Fig.~\ref{classifiers}. Four distinct feature sets served as inputs for evaluating these classifiers. The first set amalgamates all three techniques, as seen in Fig.~\ref{classifiers}(A), while the second set intertwines BP with PCA, as referenced in Fig.~\ref{classifiers}(B). The third set fuses CondEn with PCA, illustrated in Fig.~\ref{classifiers}(C), and the final set pairs PLV with the clustering coefficient, demonstrated in Fig.~\ref{classifiers}(D). Notably, linear classifiers achieved the highest accuracy, reaching 62.1\% (on Subject ZPH).

Fig.~\ref{classifiers} provides a comprehensive view of the classification accuracy results, whereas Table~\ref{tab:results} summarizes the average and standard deviation of classification performance among 12 subjects, using four different feature sets and eleven machine-learning algorithms. We noted a tangible variation in the accuracy of the classifiers across distinct methodologies and subjects in the Table. The Linear SVM consistently outperformed other algorithms, exhibiting peak accuracy of $60.03 \pm 1.72\%$ in combined features scenarios. Using the second feature set (BP + PCA) resulted in a marginal decrement in the accuracy of all classifiers, with the highest recorded at $56.73 \pm 1.80\%$ using Medium Gaussian SVM. In contrast, the third set (CondEn + PCA) enhanced accuracy for specific classifiers, with the Linear SVM being paramount, achieving $59.37 \pm 2.05\%$ at its highest. Conversely, employing the fourth set (PLV + clustering coefficient) precipitated a universal decline in overall accuracy across all classifiers, pinpointing $54.70 \pm 2.80\%$ for Linear SVM.

\section{Discussion and Conclusion}

This pilot study introduced a novel BCI pipeline that synergistically combines LLMs, particularly Generative Pre-trained Transformers (GPT-3.5 and GPT-4), and an EEG-based BCI. This is one of the first efforts to use GPT capability for this specialized intersection of neuroscience and artificial intelligence.

Eye gaze is a prominent biomarker, holding crucial information for comprehending cognitive processes in individuals involved in task-specific reading activities \cite{chen2014eye}. In this study, we conducted average fixation analyses across three distinct dimensions: on a subject-by-subject basis, concerning specific semantic relations, and at the level of individual words. We performed these analyses on data collected from 12 participants and encompassing eight different semantic relations. Our results unequivocally show that participants allocate significantly more time to words that exhibit high semantic relevance to specific relations (i.e., keywords) during inference tasks. Appendices A and B provide additional support for this observation.

Unlike traditional BCIs, which relied on precise stimulus presentation as timing markers to extract event-related EEG activities such as P300 and Steady-State Visual Evoke Potentials in well-controlled laboratory environments, our approach leveraged fixation onsets to capture EEG signals related to words during natural reading. This implementation significantly enhances the practicality of BCIs for real-world applications.

We evaluated the performance of four distinct LLMs to improve classification outcomes. Our hybrid architecture, combining GPT-3.5 and GPT-4 as word labelers with eye tracking and BCI components, demonstrated remarkable performance, achieving an impressive accuracy rate exceeding 60\% in the classification of word relevance. This enhancement was realized by applying SVMs to three domain-specific features: BP, CondEn combined with PCA, and PLV-based graph theory techniques. Carefully chose each feature for its well-established utility in BCI research and its capacity to enhance the signal-to-noise ratio. Additionally, we explored the pair-wise coherence of 5-frequency bands but ultimately decided against its use because of its computational complexity, particularly when considering the 105 EEG channels we employed.


Furthermore, we comprehensively analyzed single-word fixation statistics for 12 subjects, encompassing eight classes within the HRW and LRW groups. To account for the absence of data in eight relationship instances — Subject ZGW did not include ``JOB", ZKB lacked ``WIFE," and ZPH lacked both ``POL AFF" and ``WIFE"—we ultimately generated 184 figures ($12 \times 8 \times 2-8$), all of which are included in the supplementary materials. Our findings revealed that words within the HRW group garnered significantly higher average fixation counts than those in the LRW group. These findings provide valuable insights into how participants comprehend the reading corpus.

Despite these advances, the study has several limitations. 
This study faces challenges because of the 'black box' nature of LLMs, particularly in the context of the non-deterministic relation, such as ‘AWARD,’ where certain outputted words appear incongruous. This limitation might affect our findings' generalizability and underscore the need for a quantitative assessment to ensure the accuracy and validity of keyword identification.

Additionally, contextual complexities often influence semantic classifications. For example, ``gold" acquire distinct semantic relevance when juxtaposed with terms like ``medal." The sentences incorporating specific target terms, such as ``NATIONALITY" or ``WIFE," exhibit a significant disparity in the distribution between HRW and LRW, making them more deterministic. These discrepancies add complexity to the classification of EEG data and introduce the possibility of contamination within the dataset, especially when the meaning of words is most effectively comprehended within the context of phrases rather than in isolation.

This study underscores the potential for more expansive research on elucidating reading-related cognitive behaviors. The promise of integrating LLMs into BCIs also points towards future advancements in reading assistance technologies. While acknowledging its limitations and complexities, our work is an early yet significant contribution, paving the way for more integrated studies to foster a deeper understanding of the multifaceted interplay between neuroscience and computational linguistics.

\ifCLASSOPTIONcaptionsoff
  \newpage
\fi



%




\bibliographystyle{IEEEtran}
\bibliography{reference}

\begin{thebibliography}{10}
\providecommand{\url}[1]{#1}
\csname url@samestyle\endcsname
\providecommand{\newblock}{\relax}
\providecommand{\bibinfo}[2]{#2}
\providecommand{\BIBentrySTDinterwordspacing}{\spaceskip=0pt\relax}
\providecommand{\BIBentryALTinterwordstretchfactor}{4}
\providecommand{\BIBentryALTinterwordspacing}{\spaceskip=\fontdimen2\font plus
\BIBentryALTinterwordstretchfactor\fontdimen3\font minus
  \fontdimen4\font\relax}
\providecommand{\BIBforeignlanguage}[2]{{%
\expandafter\ifx\csname l@#1\endcsname\relax
\typeout{** WARNING: IEEEtran.bst: No hyphenation pattern has been}%
\typeout{** loaded for the language `#1'. Using the pattern for}%
\typeout{** the default language instead.}%
\else
\language=\csname l@#1\endcsname
\fi
#2}}
\providecommand{\BIBdecl}{\relax}
\BIBdecl

\bibitem{wang2023scientific}
H.~Wang, T.~Fu, Y.~Du, W.~Gao, K.~Huang, Z.~Liu, P.~Chandak, S.~Liu,
  P.~Van~Katwyk, A.~Deac \emph{et~al.}, ``Scientific discovery in the age of
  artificial intelligence,'' \emph{Nature}, vol. 620, no. 7972, pp. 47--60,
  2023.

\bibitem{singhal2023large}
K.~Singhal, S.~Azizi, T.~Tu, S.~S. Mahdavi, J.~Wei, H.~W. Chung, N.~Scales,
  A.~Tanwani, H.~Cole-Lewis, S.~Pfohl \emph{et~al.}, ``Large language models
  encode clinical knowledge,'' \emph{Nature}, pp. 1--9, 2023.

\bibitem{abdullah2022chatgpt}
M.~Abdullah, A.~Madain, and Y.~Jararweh, ``Chatgpt: Fundamentals, applications
  and social impacts,'' in \emph{2022 Ninth International Conference on Social
  Networks Analysis, Management and Security (SNAMS)}.\hskip 1em plus 0.5em
  minus 0.4em\relax IEEE, 2022, pp. 1--8.

\bibitem{bubeck2023sparks}
S.~Bubeck, V.~Chandrasekaran, R.~Eldan, J.~Gehrke, E.~Horvitz, E.~Kamar,
  P.~Lee, Y.~T. Lee, Y.~Li, S.~Lundberg \emph{et~al.}, ``Sparks of artificial
  general intelligence: Early experiments with gpt-4,'' \emph{arXiv preprint
  arXiv:2303.12712}, 2023.

\bibitem{gunning2019xai}
D.~Gunning, M.~Stefik, J.~Choi, T.~Miller, S.~Stumpf, and G.-Z. Yang,
  ``Xai—explainable artificial intelligence,'' \emph{Science robotics},
  vol.~4, no.~37, p. eaay7120, 2019.

\bibitem{just1980theory}
M.~A. Just and P.~A. Carpenter, ``A theory of reading: from eye fixations to
  comprehension.'' \emph{Psychological review}, vol.~87, no.~4, p. 329, 1980.

\bibitem{rayner1998eye}
K.~Rayner, ``Eye movements in reading and information processing: 20 years of
  research.'' \emph{Psychological bulletin}, vol. 124, no.~3, p. 372, 1998.

\bibitem{kintsch1998comprehension}
W.~Kintsch, \emph{Comprehension: A paradigm for cognition}.\hskip 1em plus
  0.5em minus 0.4em\relax Cambridge university press, 1998.

\bibitem{binz2023using}
M.~Binz and E.~Schulz, ``Using cognitive psychology to understand gpt-3,''
  \emph{Proceedings of the National Academy of Sciences}, vol. 120, no.~6, p.
  e2218523120, 2023.

\bibitem{ouyang2022training}
L.~Ouyang, J.~Wu, X.~Jiang, D.~Almeida, C.~Wainwright, P.~Mishkin, C.~Zhang,
  S.~Agarwal, K.~Slama, A.~Ray \emph{et~al.}, ``Training language models to
  follow instructions with human feedback,'' \emph{Advances in Neural
  Information Processing Systems}, vol.~35, pp. 27\,730--27\,744, 2022.

\bibitem{pandarinath2017high}
C.~Pandarinath, P.~Nuyujukian, C.~H. Blabe, B.~L. Sorice, J.~Saab, F.~R.
  Willett, L.~R. Hochberg, K.~V. Shenoy, and J.~M. Henderson, ``High
  performance communication by people with paralysis using an intracortical
  brain-computer interface,'' \emph{Elife}, vol.~6, p. e18554, 2017.

\bibitem{shawky2019towards}
D.~Shawky and A.~Badawi, ``Towards a personalized learning experience using
  reinforcement learning,'' \emph{Machine learning paradigms: Theory and
  application}, pp. 169--187, 2019.

\bibitem{kutas2011thirty}
M.~Kutas and K.~D. Federmeier, ``Thirty years and counting: finding meaning in
  the n400 component of the event-related brain potential (erp),'' \emph{Annual
  review of psychology}, vol.~62, pp. 621--647, 2011.

\bibitem{kutas1980reading}
M.~Kutas and S.~A. Hillyard, ``Reading senseless sentences: Brain potentials
  reflect semantic incongruity,'' \emph{Science}, vol. 207, no. 4427, pp.
  203--205, 1980.

\bibitem{kintsch1988role}
W.~Kintsch, ``The role of knowledge in discourse comprehension: a
  construction-integration model.'' \emph{Psychological review}, vol.~95,
  no.~2, p. 163, 1988.

\bibitem{evans2008dual}
J.~S.~B. Evans, ``Dual-processing accounts of reasoning, judgment, and social
  cognition,'' \emph{Annu. Rev. Psychol.}, vol.~59, pp. 255--278, 2008.

\bibitem{rumelhart2017schemata}
D.~E. Rumelhart, ``Schemata: The building blocks of cognition,'' in
  \emph{Theoretical issues in reading comprehension}.\hskip 1em plus 0.5em
  minus 0.4em\relax Routledge, 2017, pp. 33--58.

\bibitem{anderson2018role}
R.~C. Anderson, ``Role of the reader's schema in comprehension, learning, and
  memory,'' in \emph{Theoretical models and processes of literacy}.\hskip 1em
  plus 0.5em minus 0.4em\relax Routledge, 2018, pp. 136--145.

\bibitem{johnson1983mental}
P.~N. Johnson-Laird, \emph{Mental models: Towards a cognitive science of
  language, inference, and consciousness}.\hskip 1em plus 0.5em minus
  0.4em\relax Harvard University Press, 1983, no.~6.

\bibitem{mcnamara2009toward}
D.~S. McNamara and J.~Magliano, ``Toward a comprehensive model of
  comprehension,'' \emph{Psychology of learning and motivation}, vol.~51, pp.
  297--384, 2009.

\bibitem{baretta2012investigating}
L.~Baretta, L.~M.~B. Tomitch, V.~K. Lim, and K.~E. Waldie, ``Investigating
  reading comprehension through eeg,'' \emph{Ilha do Desterro: A Journal of
  English Language, Literatures in English and Cultural Studies}, no.~63, pp.
  69--99, 2012.

\bibitem{bornstein2011cognitive}
M.~H. Bornstein and M.~E. Lamb, \emph{Cognitive development: An advanced
  textbook}.\hskip 1em plus 0.5em minus 0.4em\relax Taylor \& Francis, 2011.

\bibitem{mridha2021brain}
M.~F. Mridha, S.~C. Das, M.~M. Kabir, A.~A. Lima, M.~R. Islam, and Y.~Watanobe,
  ``Brain-computer interface: Advancement and challenges,'' \emph{Sensors},
  vol.~21, no.~17, p. 5746, 2021.

\bibitem{li2021reading}
Q.~Li, \emph{Reading Comprehension Analysis and Prediction Based on EEG and
  Eye-Tracking Techniques}.\hskip 1em plus 0.5em minus 0.4em\relax University
  of California, San Diego, 2021.

\bibitem{zeng2018eeg}
H.~Zeng, C.~Yang, G.~Dai, F.~Qin, J.~Zhang, and W.~Kong, ``Eeg classification
  of driver mental states by deep learning,'' \emph{Cognitive neurodynamics},
  vol.~12, pp. 597--606, 2018.

\bibitem{seitz2008valuating}
R.~J. Seitz, R.~Sch{\"a}fer, D.~Scherfeld, S.~Friederichs, K.~Popp, H.-J.
  Wittsack, N.~Azari, and M.~Franz, ``Valuating other people’s emotional face
  expression: a combined functional magnetic resonance imaging and
  electroencephalography study,'' \emph{Neuroscience}, vol. 152, no.~3, pp.
  713--722, 2008.

\bibitem{tanaka2014neural}
M.~Tanaka, A.~Ishii, and Y.~Watanabe, ``Neural effects of mental fatigue caused
  by continuous attention load: a magnetoencephalography study,'' \emph{Brain
  research}, vol. 1561, pp. 60--66, 2014.

\bibitem{jenkins2019rethinking}
A.~C. Jenkins, ``Rethinking cognitive load: a default-mode network
  perspective,'' \emph{Trends in Cognitive Sciences}, vol.~23, no.~7, pp.
  531--533, 2019.

\bibitem{wang2014eye}
Q.~Wang, S.~Yang, M.~Liu, Z.~Cao, and Q.~Ma, ``An eye-tracking study of website
  complexity from cognitive load perspective,'' \emph{Decision support
  systems}, vol.~62, pp. 1--10, 2014.

\bibitem{hollenstein2018ZuCo}
N.~Hollenstein, J.~Rotsztejn, M.~Troendle, A.~Pedroni, C.~Zhang, and N.~Langer,
  ``Zuco, a simultaneous eeg and eye-tracking resource for natural sentence
  reading,'' \emph{Scientific data}, vol.~5, no.~1, pp. 1--13, 2018.

\bibitem{brouwer2012getting}
H.~Brouwer, H.~Fitz, and J.~Hoeks, ``Getting real about semantic illusions:
  Rethinking the functional role of the p600 in language comprehension,''
  \emph{Brain research}, vol. 1446, pp. 127--143, 2012.

\bibitem{manning2014stanford}
C.~D. Manning, M.~Surdeanu, J.~Bauer, J.~R. Finkel, S.~Bethard, and
  D.~McClosky, ``The stanford corenlp natural language processing toolkit,'' in
  \emph{Proceedings of 52nd annual meeting of the association for computational
  linguistics: system demonstrations}, 2014, pp. 55--60.

\bibitem{aydore2013note}
S.~Aydore, D.~Pantazis, and R.~M. Leahy, ``A note on the phase locking value
  and its properties,'' \emph{Neuroimage}, vol.~74, pp. 231--244, 2013.

\bibitem{NIPS1998_0d4f4805}
T.-P. Jung, S.~Makeig, M.~Westerfield, J.~Townsend, E.~Courchesne, and T.~J.
  Sejnowski, ``Analyzing and visualizing single-trial event-related
  potentials,'' in \emph{Advances in Neural Information Processing Systems},
  M.~Kearns, S.~Solla, and D.~Cohn, Eds., vol.~11.\hskip 1em plus 0.5em minus
  0.4em\relax MIT Press, 1998.

\bibitem{rubinov2010complex}
M.~Rubinov and O.~Sporns, ``Complex network measures of brain connectivity:
  uses and interpretations,'' \emph{Neuroimage}, vol.~52, no.~3, pp.
  1059--1069, 2010.

\bibitem{zhang2021emergency}
Y.~Zhang, Y.~Liao, Y.~Zhang, and L.~Huang, ``Emergency braking intention detect
  system based on k-order propagation number algorithm: a network
  perspective,'' \emph{Brain Sciences}, vol.~11, no.~11, p. 1424, 2021.

\bibitem{ding2022using}
W.~Ding, Y.~Zhang, and L.~Huang, ``Using a novel functional brain network
  approach to locate important nodes for working memory tasks,''
  \emph{International journal of environmental research and public health},
  vol.~19, no.~6, p. 3564, 2022.

\bibitem{chen2022research}
Y.~Chen, Y.~Zhang, W.~Ding, F.~Cui, and L.~Huang, ``Research on working memory
  states based on weighted-order propagation number algorithm: An eeg
  perspective,'' \emph{Journal of Sensors}, vol. 2022, 2022.

\bibitem{fornito2016fundamentals}
A.~Fornito, A.~Zalesky, and E.~Bullmore, \emph{Fundamentals of brain network
  analysis}.\hskip 1em plus 0.5em minus 0.4em\relax Academic press, 2016.

\bibitem{bullmore2009complex}
E.~Bullmore and O.~Sporns, ``Complex brain networks: graph theoretical analysis
  of structural and functional systems,'' \emph{Nature reviews neuroscience},
  vol.~10, no.~3, pp. 186--198, 2009.

\bibitem{chiang2023using}
K.-J. Chiang, S.~Dong, C.-K. Cheng, and T.-P. Jung, ``Using eeg signals to
  assess workload during memory retrieval in a real-world scenario,''
  \emph{Journal of Neural Engineering}, vol.~20, no.~3, p. 036010, 2023.

\bibitem{Lotte_2018}
\BIBentryALTinterwordspacing
F.~Lotte, L.~Bougrain, A.~Cichocki, M.~Clerc, M.~Congedo, A.~Rakotomamonjy, and
  F.~Yger, ``A review of classification algorithms for eeg-based
  brain–computer interfaces: a 10 year update,'' \emph{Journal of Neural
  Engineering}, vol.~15, no.~3, p. 031005, apr 2018. [Online]. Available:
  \url{https://dx.doi.org/10.1088/1741-2552/aab2f2}
\BIBentrySTDinterwordspacing

\bibitem{Lawhern_2018}
\BIBentryALTinterwordspacing
V.~J. Lawhern, A.~J. Solon, N.~R. Waytowich, S.~M. Gordon, C.~P. Hung, and
  B.~J. Lance, vol.~15, no.~5, p. 056013, jul 2018. [Online]. Available:
  \url{https://dx.doi.org/10.1088/1741-2552/aace8c}
\BIBentrySTDinterwordspacing

\bibitem{Craik_2019}
\BIBentryALTinterwordspacing
A.~Craik, Y.~He, and J.~L. Contreras-Vidal, ``Deep learning for
  electroencephalogram (eeg) classification tasks: a review,'' \emph{Journal of
  Neural Engineering}, vol.~16, no.~3, p. 031001, apr 2019. [Online].
  Available: \url{https://dx.doi.org/10.1088/1741-2552/ab0ab5}
\BIBentrySTDinterwordspacing

\bibitem{li2022wearablebased}
Y.~Li, R.~Yin, H.~Park, Y.~Kim, and P.~Panda, ``Wearable-based human activity
  recognition with spatio-temporal spiking neural networks,'' 2022.

\bibitem{10.1093/cercor/bhh167}
\BIBentryALTinterwordspacing
C.~Tallon-Baudry, O.~Bertrand, M.-A. Hénaff, J.~Isnard, and C.~Fischer,
  ``{Attention Modulates Gamma-band Oscillations Differently in the Human
  Lateral Occipital Cortex and Fusiform Gyrus},'' \emph{Cerebral Cortex},
  vol.~15, no.~5, pp. 654--662, 09 2004. [Online]. Available:
  \url{https://doi.org/10.1093/cercor/bhh167}
\BIBentrySTDinterwordspacing

\bibitem{wyart2009ongoing}
V.~Wyart and C.~Tallon-Baudry, ``How ongoing fluctuations in human visual
  cortex predict perceptual awareness: baseline shift versus decision bias,''
  \emph{Journal of Neuroscience}, vol.~29, no.~27, pp. 8715--8725, 2009.

\bibitem{palva2011localization}
S.~Palva, S.~Kulashekhar, M.~H{\"a}m{\"a}l{\"a}inen, and J.~M. Palva,
  ``Localization of cortical phase and amplitude dynamics during visual working
  memory encoding and retention,'' \emph{Journal of Neuroscience}, vol.~31,
  no.~13, pp. 5013--5025, 2011.

\bibitem{chen2014eye}
S.-C. Chen, H.-C. She, M.-H. Chuang, J.-Y. Wu, J.-L. Tsai, and T.-P. Jung,
  ``Eye movements predict students' computer-based assessment performance of
  physics concepts in different presentation modalities,'' \emph{Computers \&
  Education}, vol.~74, pp. 61--72, 2014.

\end{thebibliography}

%










\end{document}



\section*{Appendix A: Relation-wise Average Fixation}

\begin{figure}[h]
    \centering
    \begin{subfigure}[b]{0.49\textwidth}
        \includegraphics[width=\textwidth]{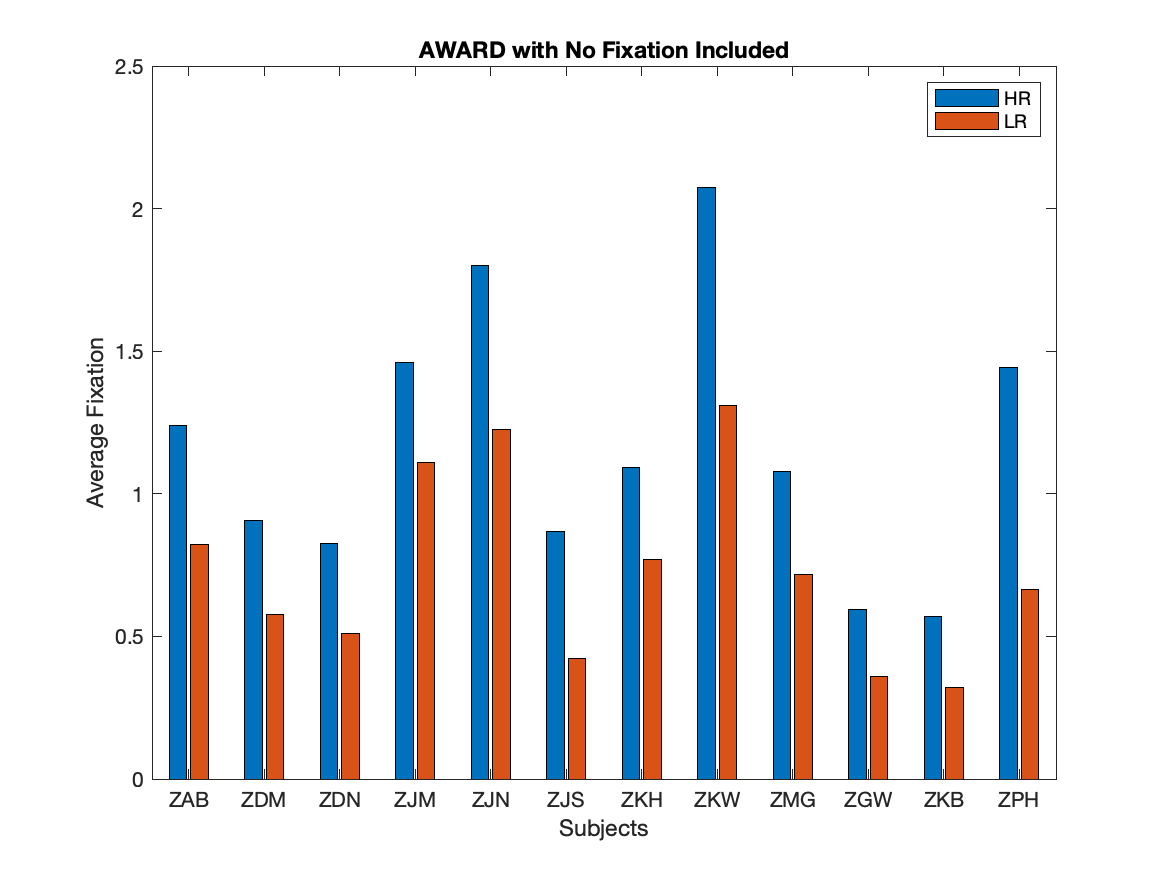}
        \caption{AWARD with No Fixation Included}
    \end{subfigure}
    \hfill
    \begin{subfigure}[b]{0.49\textwidth}
        \includegraphics[width=\textwidth]{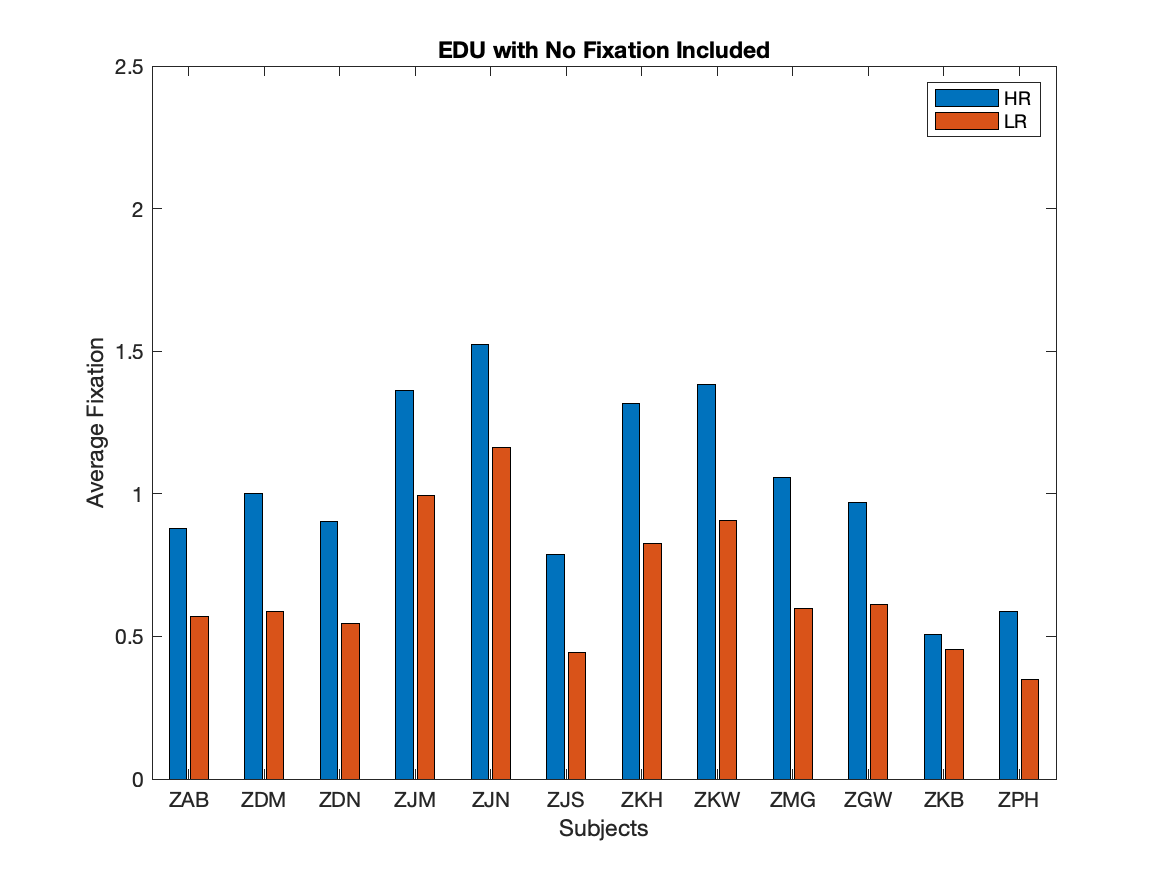}
        \caption{EDU with No Fixation Included}
    \end{subfigure}

    \vspace{1em}  

    \begin{subfigure}[b]{0.49\textwidth}
        \includegraphics[width=\textwidth]{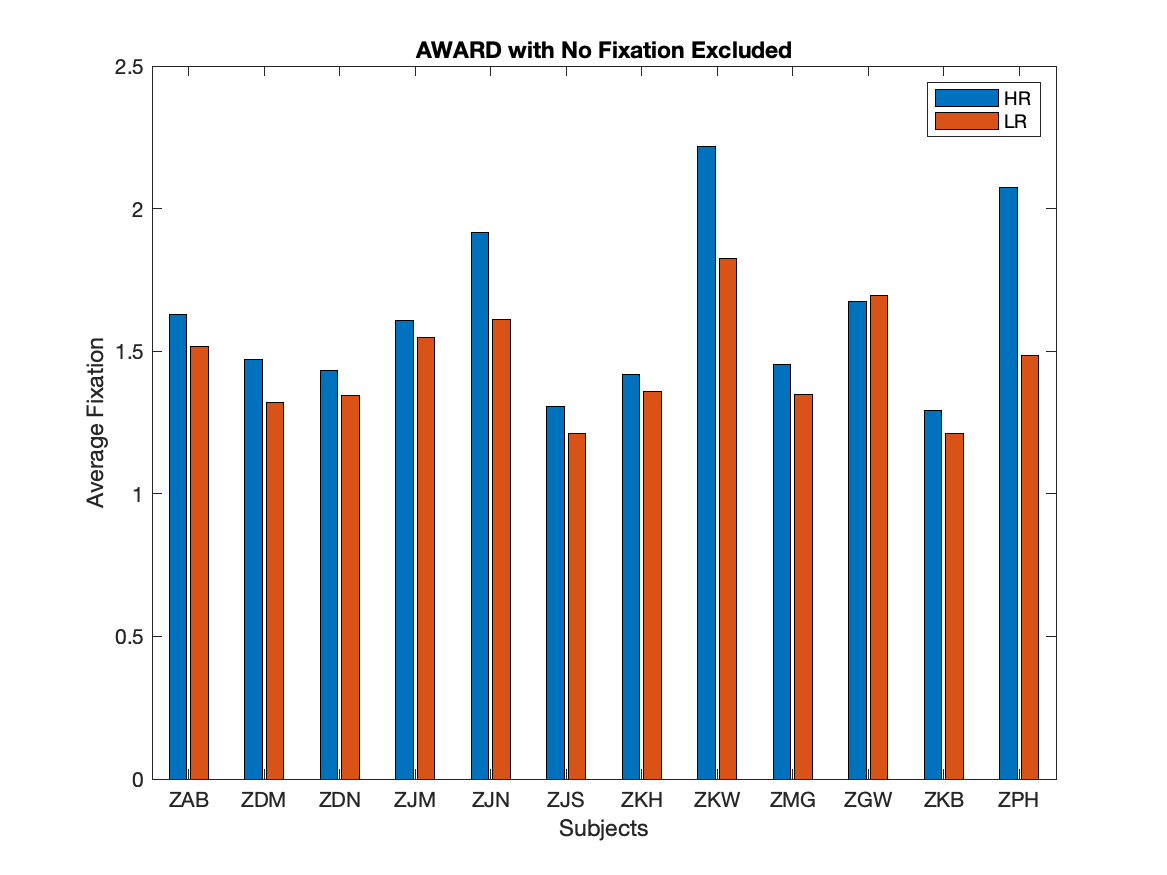}
        \caption{AWARD with No Fixation Excluded}
    \end{subfigure}
    \hfill
    \begin{subfigure}[b]{0.49\textwidth}
        \includegraphics[width=\textwidth]{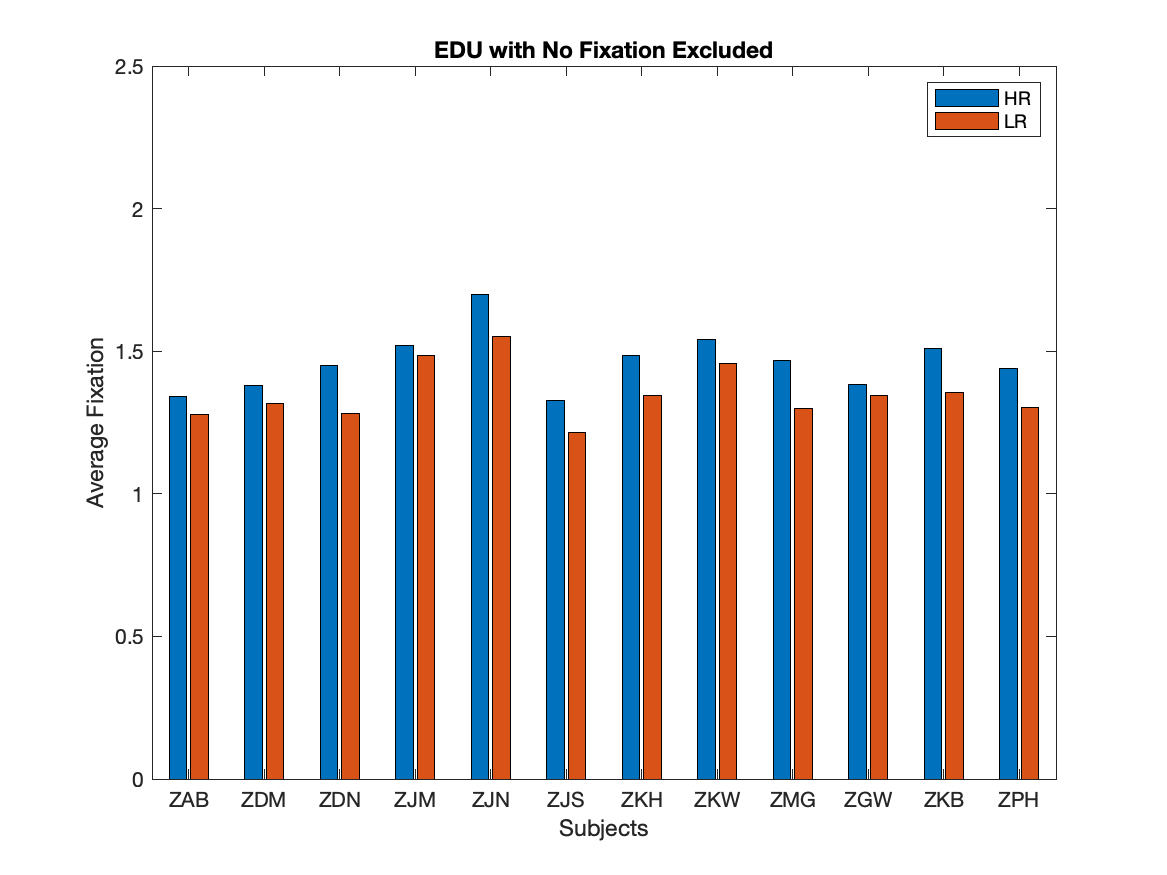}
        \caption{EDU with No Fixation Excluded}
    \end{subfigure}
    
    \caption{Supplementary figures showing the average fixation for AWARD and EDUCATION, both with and without fixation included}
\end{figure}

\section*{Appendix B: Word-wise Average Fixation}

\includegraphics[width=0.45\textwidth]{SupplementaryMaterialFigs/ZAB Fixation v. High-Related Words to AWARD.png}
\hfill
\includegraphics[width=0.45\textwidth]{SupplementaryMaterialFigs/ZAB Fixation v. Low-Related Words to AWARD.png}

\captionof{figure}{Subject ZAB's Fixation in AWARD Relation Category}
\vspace{1em} 

\includegraphics[width=0.45\textwidth]{SupplementaryMaterialFigs/ZAB Fixation v. High-Related Words to EDUCATION.png}
\hfill
\includegraphics[width=0.45\textwidth]{SupplementaryMaterialFigs/ZAB Fixation v. Low-Related Words to EDUCATION.png}

\captionof{figure}{Subject ZAB's Fixation in EDUCATION Relation Category}
\vspace{1em} 

\includegraphics[width=0.45\textwidth]{SupplementaryMaterialFigs/ZAB Fixation v. High-Related Words to EMPLOYER.png}
\hfill
\includegraphics[width=0.45\textwidth]{SupplementaryMaterialFigs/ZAB Fixation v. Low-Related Words to EMPLOYER.png}

\captionof{figure}{Subject ZAB's Fixation in EMPLOYER Relation Category}
\vspace{1em} 

\includegraphics[width=0.45\textwidth]{SupplementaryMaterialFigs/ZAB Fixation v. High-Related Words to FOUNDER.png}
\hfill
\includegraphics[width=0.45\textwidth]{SupplementaryMaterialFigs/ZAB Fixation v. Low-Related Words to FOUNDER.png}

\captionof{figure}{Subject ZAB's Fixation in FOUNDER Relation Category}
\vspace{1em} 

\includegraphics[width=0.45\textwidth]{SupplementaryMaterialFigs/ZAB Fixation v. High-Related Words to JOB TITLE.png}
\hfill
\includegraphics[width=0.45\textwidth]{SupplementaryMaterialFigs/ZAB Fixation v. Low-Related Words to JOB TITLE.png}

\captionof{figure}{Subject ZAB's Fixation in JOB TITLE Relation Category}
\vspace{1em} 

\includegraphics[width=0.45\textwidth]{SupplementaryMaterialFigs/ZAB Fixation v. High-Related Words to NATIONALITY.png}
\hfill
\includegraphics[width=0.45\textwidth]{SupplementaryMaterialFigs/ZAB Fixation v. Low-Related Words to NATIONALITY.png}

\captionof{figure}{Subject ZAB's Fixation in NATIONALITY Relation Category}
\vspace{1em} 

\includegraphics[width=0.45\textwidth]{SupplementaryMaterialFigs/ZAB Fixation v. High-Related Words to POLITICAL AFFILIATION.png}
\hfill
\includegraphics[width=0.45\textwidth]{SupplementaryMaterialFigs/ZAB Fixation v. Low-Related Words to POLITICAL AFFILIATION.png}

\captionof{figure}{Subject ZAB's Fixation in POLITICAL AFFILIATION Relation Category}
\vspace{1em} 

\includegraphics[width=0.45\textwidth]{SupplementaryMaterialFigs/ZAB Fixation v. High-Related Words to WIFE.png}
\hfill
\includegraphics[width=0.45\textwidth]{SupplementaryMaterialFigs/ZAB Fixation v. Low-Related Words to WIFE.png}

\captionof{figure}{Subject ZAB's Fixation in WIFE Relation Category}
\vspace{1em} 

\includegraphics[width=0.45\textwidth]{SupplementaryMaterialFigs/ZDM Fixation v. High-Related Words to AWARD.png}
\hfill
\includegraphics[width=0.45\textwidth]{SupplementaryMaterialFigs/ZDM Fixation v. Low-Related Words to AWARD.png}

\captionof{figure}{Subject ZDM's Fixation in AWARD Relation Category}
\vspace{1em} 

\includegraphics[width=0.45\textwidth]{SupplementaryMaterialFigs/ZDM Fixation v. High-Related Words to EDUCATION.png}
\hfill
\includegraphics[width=0.45\textwidth]{SupplementaryMaterialFigs/ZDM Fixation v. Low-Related Words to EDUCATION.png}

\captionof{figure}{Subject ZDM's Fixation in EDUCATION Relation Category}
\vspace{1em} 

\includegraphics[width=0.45\textwidth]{SupplementaryMaterialFigs/ZDM Fixation v. High-Related Words to EMPLOYER.png}
\hfill
\includegraphics[width=0.45\textwidth]{SupplementaryMaterialFigs/ZDM Fixation v. Low-Related Words to EMPLOYER.png}

\captionof{figure}{Subject ZDM's Fixation in EMPLOYER Relation Category}
\vspace{1em} 

\includegraphics[width=0.45\textwidth]{SupplementaryMaterialFigs/ZDM Fixation v. High-Related Words to FOUNDER.png}
\hfill
\includegraphics[width=0.45\textwidth]{SupplementaryMaterialFigs/ZDM Fixation v. Low-Related Words to FOUNDER.png}

\captionof{figure}{Subject ZDM's Fixation in FOUNDER Relation Category}
\vspace{1em} 

\includegraphics[width=0.45\textwidth]{SupplementaryMaterialFigs/ZDM Fixation v. High-Related Words to JOB TITLE.png}
\hfill
\includegraphics[width=0.45\textwidth]{SupplementaryMaterialFigs/ZDM Fixation v. Low-Related Words to JOB TITLE.png}

\captionof{figure}{Subject ZDM's Fixation in JOB TITLE Relation Category}
\vspace{1em} 

\includegraphics[width=0.45\textwidth]{SupplementaryMaterialFigs/ZDM Fixation v. High-Related Words to NATIONALITY.png}
\hfill
\includegraphics[width=0.45\textwidth]{SupplementaryMaterialFigs/ZDM Fixation v. Low-Related Words to NATIONALITY.png}

\captionof{figure}{Subject ZDM's Fixation in NATIONALITY Relation Category}
\vspace{1em} 

\includegraphics[width=0.45\textwidth]{SupplementaryMaterialFigs/ZDM Fixation v. High-Related Words to POLITICAL AFFILIATION.png}
\hfill
\includegraphics[width=0.45\textwidth]{SupplementaryMaterialFigs/ZDM Fixation v. Low-Related Words to POLITICAL AFFILIATION.png}

\captionof{figure}{Subject ZDM's Fixation in POLITICAL AFFILIATION Relation Category}
\vspace{1em} 

\includegraphics[width=0.45\textwidth]{SupplementaryMaterialFigs/ZDM Fixation v. High-Related Words to WIFE.png}
\hfill
\includegraphics[width=0.45\textwidth]{SupplementaryMaterialFigs/ZDM Fixation v. Low-Related Words to WIFE.png}

\captionof{figure}{Subject ZDM's Fixation in WIFE Relation Category}
\vspace{1em} 